\documentclass{article}

\usepackage{graphicx}
\usepackage{hyperref}
\usepackage{amsmath}
\usepackage{amsfonts}
\usepackage{ upgreek }
\usepackage{array} 
\usepackage{longtable}
\usepackage{pdflscape}
\usepackage{afterpage}
\usepackage{float}
\usepackage{capt-of}
\usepackage{geometry} 
\usepackage{color}

\def\eg{e.g.,\/}
\def\ie{i.e.,\/}
\def\etal{et al.\/}

\makeatletter
\def\ps@pprintTitle{%
  \let\@oddhead\@empty
  \let\@evenhead\@empty
  \let\@oddfoot\@empty
  \let\@evenfoot\@oddfoot
}
\makeatother

\newcolumntype{L}[1]{>{\raggedright\let\newline\\\arraybackslash\hspace{0pt}}m{#1}}
\newcolumntype{C}[1]{>{\centering\let\newline\\\arraybackslash\hspace{0pt}}m{#1}}
\newcolumntype{R}[1]{>{\raggedleft\let\newline\\\arraybackslash\hspace{0pt}}m{#1}}

\newcommand\nPapers{77}
\newcommand\nScansPerStudy{59}
\newcommand\nPapersUsedAtlases{48}
\newcommand\nPapersUsedLandmarks{6}
\newcommand\nPredictionPapers{67}
\newcommand\nPapersUsedUnsupervised{Fifteen}
\newcommand\nRegressionPapers{2}
\newcommand\nClassificationPapers{65}
\newcommand\nTwoClassPapers{55}
\newcommand\nMultiClassPapers{10}
\newcommand\nPapersUsedSVM{47}
\newcommand\nEdgeFeaturePapers{38}
\newcommand\nImbalancedPapers{18}
\newcommand\nPapersUsedAcc{64}
\newcommand\nPapersUseLOO{40}
\newcommand\nPapersUsedLKO{6}

\begin{document}

\title{Machine Learning on Human Connectome Data from MRI}
\author{Colin J Brown and Ghassan Hamarneh \\ Medical Image Analysis Lab, Simon Fraser University, Burnaby, BC, Canada}
\date{}
\maketitle

\begin{abstract}
Functional MRI (fMRI) and diffusion MRI (dMRI) are non-invasive imaging modalities that allow \emph{in-vivo} analysis of a patient's brain network (known as a connectome). Use of these technologies has enabled faster and better diagnoses and treatments of neurological disorders and a deeper understanding of the human brain. Recently, researchers have been exploring the application of machine learning models to connectome data in order to predict clinical outcomes and analyze the importance of subnetworks in the brain. Connectome data has unique properties, which present both special challenges and opportunities when used for machine learning. The purpose of this work is to review the literature on the topic of applying machine learning models to MRI-based connectome data. This field is growing rapidly and now encompasses a large body of research. To summarize the research done to date, we provide a comparative, structured summary of \nPapers~relevant works, tabulated according to different criteria, that represent the majority of the literature on this topic. (We also published a living version of this table online at \href{http://connectomelearning.cs.sfu.ca/}{http://connectomelearning.cs.sfu.ca/} that the community can continue to contribute to.)  After giving an overview of how connectomes are constructed from dMRI and fMRI data, we discuss the variety of machine learning tasks that have been explored with connectome data. We then compare the advantages and drawbacks of different machine learning approaches that have been employed, discussing different feature selection and feature extraction schemes, as well as the learning models and regularization penalties themselves. Throughout this discussion, we focus particularly on how the methods are adapted to the unique nature of graphical connectome data. Finally, we conclude by summarizing the current state of the art and by outlining what we believe are strategic directions for future research.
\end{abstract}


\clearpage 


\begin{table}[H]
\centering
\caption{List of common mathmatical symbols used in this paper.}
\renewcommand{\tabcolsep}{4pt}
\begin{tabular}{| c l |}
\hline
Symbol & Definition\\
\hline
$N$ & number of scans in a dataset \\
$M$ & number of features \\
$G(E,V)$ & graph (\ie\ brain network or connectome) \\
$E$ & set of edges \\
$V$ & set of vertices \\
$A$ & adjacency matrix of a connectome \\
$L(\cdot)$ & loss function (\ie\ data term) \\
$R(\cdot)$ & regularization function \\
$f(\cdot)$ & learned prediction function \\
$\mathbf{y}$ & vector of instance labels (can be continuous or categorical)\\
$\hat{\mathbf{y}}$ & vector of predicted labels\\
$X$ & matrix of feature row vectors\\
$\Theta$ & set of model parameters (to be learned)\\
$\mathbf{w}$ & vector of linear model weights \\
$w_{ij}$ & feature at edge between nodes $i$ and $j$ \\
$\lambda$ & hyper-parameter \\
$\mathcal{L}$ & graph Laplacian \\
$\mathcal{N}(i)$ & set of nodes neighbouring node $i$ \\
$d_i$ & degree of node $i$ \\
$P_{j,k}$ & path (sequence) of edges between nodes $j$ and $k$ \\
$\mathcal{O}(\cdot)$ & big-O notation \\
$\mathbb{R}$ & set of real numbers \\
\hline
\end{tabular}
\label{tab:notation}
\end{table}

\clearpage 

\section{Introduction}
The human brain is a complex network of neurons~\cite{Sporns2005}. Advances in imaging technologies such as functional MRI (fMRI) and diffusion MRI (dMRI) that allow non-invasive, \emph{in-vivo} analysis of a patient's brain network (known as a connectome), along with increased access to these technologies by researchers and clinicians, have substantially enhanced our understanding of the human brain, improving diagnoses and treatments of neurological disorders~\cite{bassett2009,nimsky2005preoperative,Henderson2012}. Over the past few years, researchers have been starting to apply the tools of machine learning to connectome data in order to perform tasks such as prediction of clinical outcomes and analysis of subnetworks in the brain. Here, we first outline the unique characteristics of connectome data (Section~\ref{sec:challengesOpportunities}) and then outline the purpose, scope and layout of this review (Section~\ref{sec:scope}).

\subsection{Special Challenges and Opportunities of Connectome Data}
\label{sec:challengesOpportunities}
Connectome data is unique in a variety of ways which both pose unique challenges and provide unique opportunities when used in the context of machine learning. Whereas standard medical images are grid-like with each pixel neighbouring only other pixels that are spatially near-by, connectomes have a more general topology. Each brain region (represented as a node in the network) may be connected, either structurally and/or functionally, to any other brain region (with each connection between two regions respresented as a binary or weighted edge in the network). This structure conveys intrinsic information about the connectivity of the brain that is not explicitly represented in other modalities or image formats. This topological information has been found to be useful as a discriminative biomarker of a variety of neurological conditions~\cite{bassett2009,Ball2015,Nir2012,Jahanshad2015,Rudie2013,Sun2015,Han2014}. Nevertheless, many types of features that have successfully been applied to extract information from grid-like image data (\eg\ scale invariant feature transform features~\cite{Lowe1999}, histogram of oriented gradient features~\cite{Dalal2010} and learned convolutional neural network (CNN) features~\cite{Krizhevsky2012}) are inapplicable or would need to be modified for use on network structured data. The adoption of connectome data has required new kinds of features to be explored (Section \ref{sec:connectomeFeatures}).

Another aspect of connectome data that differs from grid-like image data is that, in most cases, connectome nodes have intrinsic correspondence between subjects (and between multiple scans of the same subject). Thus, once the connectomes are constructed, each edge and node has a particular biological interpretation and comparison between nodes or edges of multiple connectomes is trivial. In contrast, the biological interpretation of individual voxels in an image depend completely on the frame of reference and two images must undergo image registration in order to establish correspondence. Nonetheless, constructing a connectome from raw image data is typically a multi-stage pipeline (Section~\ref{sec:connectomeConstruction}) which can introduce error and noise at each step~\cite{Zhong2015,Smith2011}.

The long acquisition time and high cost of acquiring dMRI and fMRI scans means that many studies are performed and validated over relatively few scans. For instance, examining the studies listed in Table~\ref{tab:depthReportTable} shows that, while there is a trend of connectome datasets becoming larger (see Fig.~\ref{fig:numScansVsYear}), the median number of scans used in each study is only $N=\nScansPerStudy$. Though connectomes are typically lower dimensional than grid-like images (\eg\ thousands of connections versus millions of voxels), this dimensionality is nearly always larger than the number of scans. Of the papers covered in this review, the median number of basic edge features in each connectome was $M=2850$, giving rise to the cannonical problem of $N \ll M$~\cite{Mwangi2014}, known as the high dimensional small sample size (HDSSS) problem. In order to prevent model learning from being an ill-posed problem~\cite{Bertero1988}, feature selection, data augmentation, dimensionality reduction (Section~\ref{sec:featureSelection}) and model regularization ( Section~\ref{sec:learningModels}) are often essential components of a machine learning pipeline for connectome data.

Another unique aspect of machine learning on connectome data is the biology of the brain itself: The brain is a highly complex organ comprising billions of neurons~\cite{Pakkenberg1997} and the structural arrangement of these neurons changes greatly over a persons lifetime~\cite{Collin2013}. Also, the connectivity of the brain is altered by learning and expierences~\cite{Draganski2008}, as well as injury and pathology, so the exact structure of each person's brain is inherently unique, at least on fine scales. Thus, modeling variation in connectivity of the brain over a population may be especially challenging, perhaps requiring flexible and highly non-linear models~\cite{Mitra2016}. Furthermore, ground truth class labels for images of the brain often rely on measurements or tests of ability which may introduce subjective bias from the rater. Thus, connectome data often have noisy labels as well as noisy features.\\

\subsection{Purpose, Related Works and Scope}
\label{sec:scope}
The purpose of this work is to provide a thorough review of studies that have applied machine learning models to MRI-based connectome data. To do this, we have created a table of \nPapers~works (Table~\ref{tab:depthReportTable}) that represent the majority of the literature on this topic. The field is growing rapidly and now encompasses a large body of research. Fig.~\ref{fig:numPapersVsYear} shows the number of studies on machine learning with connectome data published per year (as listed in Table~\ref{tab:depthReportTable}). No papers found were published prior to 2009 but the average number of publications has increased nearly every year since. Note that the entry for 2016 represents only those papers published prior to September and so, while the count for 2016 seems low, especially for mutli-modal studies, we extrapolate that it will outgrow the tally for 2015 by the end of the year.

Performing a search using Google Scholar\footnote{\url{scholar.google.com}} with the search pattern ``machine learning'' + (connectome OR ``brain network'') + (fMRI OR dMRI OR DTI OR ``diffusion MRI'' OR ``functional MRI'' OR ``diffusion tensor'') returns 2270 results but, when restricted to only return results from before 2009, the same search only returns 110 results (as of September, 2016). Fig.~\ref{fig:numResultsVsYear} shows this trend over time. While the results of these searches likely overestimate the true number of studies on this topic (\eg\ due to finding the keywords out of context or in references to other works), it suggests that the relative popularity of studying machine learning on connectome data is increasing. Thus, there is a present need for surveys on machine learning with connectome data that summarize the methods and results of the research that has been done in the past decade or so. \\

\begin{minipage}{0.48\textwidth }
\includegraphics[width=0.99\textwidth]{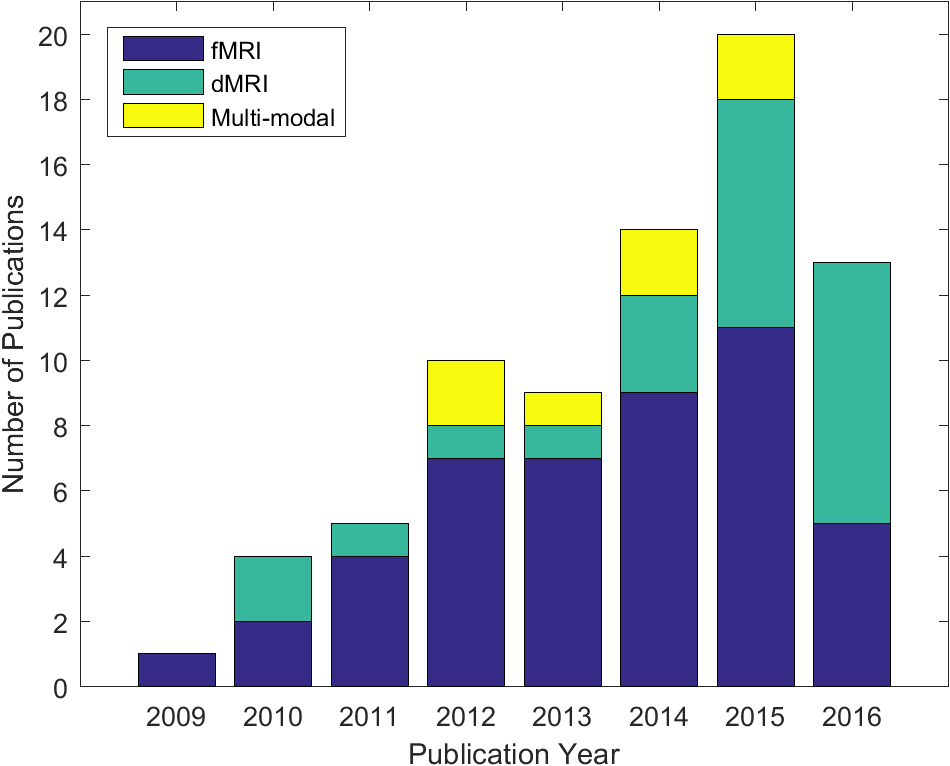}
\captionof{figure}{Number of studies (found by this survey) on the topic of machine learning for MRI-based connectome data versus year of publication. Colors represent the number of fMRI (blue), dMRI (green) and mutli-modality (yellow) studies by year. Note that for papers which studied multiple datasets independently, each dataset is counted separately in this chart.}
\label{fig:numPapersVsYear}
\end{minipage}
\begin{minipage}{.02\textwidth }
~
\end{minipage}
\begin{minipage}{.48\textwidth }
\includegraphics[width=0.99\textwidth]{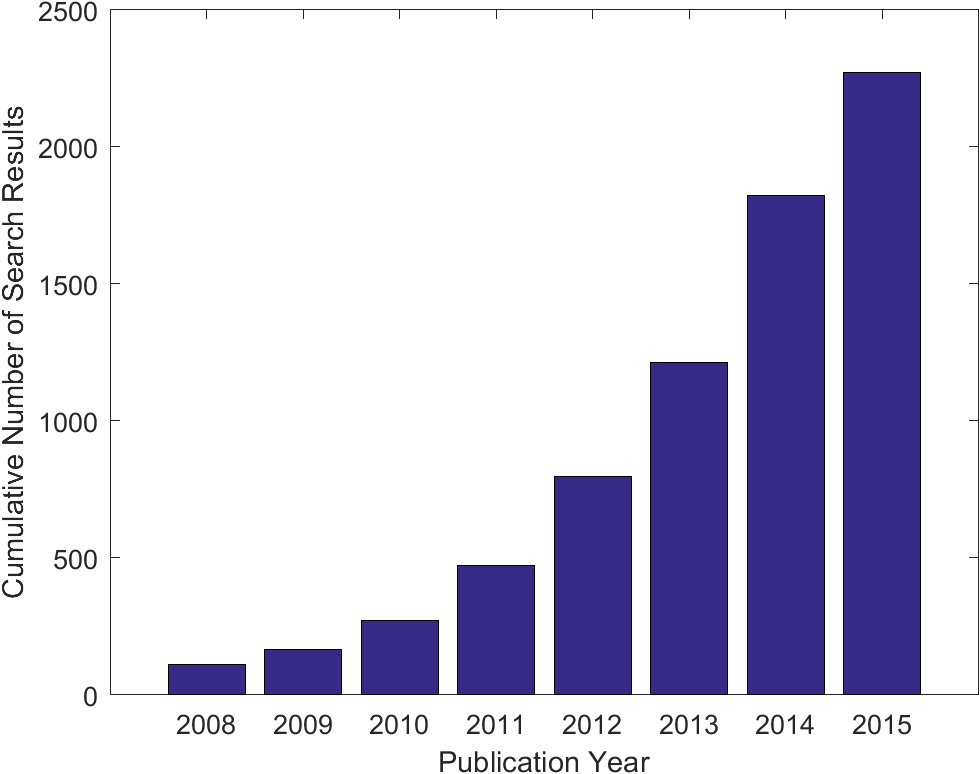}
\captionof{figure}{Number of search results found by Google Scholar for search pattern ``machine learning'' + (connectome OR ``brain network'') + (fMRI OR dMRI OR DTI  OR ``diffusion MRI'' OR ``functional MRI'' OR ``diffusion tensor'') versus maximum publication year. Note that the results are cumulative and that these searches likely greatly overestimate the actual number of publications on the topic but do suggest a relative increase in scientific interest on the topic over time.}
\label{fig:numResultsVsYear}
\end{minipage}\\

To date, few surveys on applying machine learning to connectome data have been written. Richiardi \etal\ wrote a survey on this topic in 2013~\cite{Richiardi2013} however it only covered machine learning on fMRI based connectome data. Furthermore, since this paper was published a great deal of work has been done (\ie\ Fig.~\ref{fig:numPapersVsYear} shows that more than half of the papers listed in the present review were published after 2013). In 2014, Varoquaux and Thirion published a review that was also focused on functional connectome data (including EEG) but did not perform an in-depth comparitive analysis of relevant works as we do here~\cite{Varoquaux2014}. Mwangi \etal's review (2014) examined papers across all neuroimaging modalities but focused only on feature selection and dimensionality reduction techniques~\cite{Mwangi2014}. Bassett and Lynall's survey (2013) covered a variety of network based approaches for analysis of the brain but didn't focus specifically on machine learning~\cite{Bassett2013}.

The scope of the papers included in Table~\ref{tab:depthReportTable} is limited specifically to those that applied machine learning to MRI derived connectome data. While many studies have performed statistical analsysis studies (\eg\ group difference studies) on functional and structural connectome data (\eg\ \cite{bassett2009,Brown2014,Rudie2013,Taylor2015}), papers were only included in this review if they employed supervised or unsupervised machine learning models. In the case of supervised learning, we only examined studies where the models were trained and tested on distinct datasets or studies that used a cross-validation scheme. Because the focus of this review is on connectome data from MRI images, papers that primarily examined electro-encephalography (EEG) data (\eg\ Micheloyannis \etal~\cite{Micheloyannis2006}), magneto-encephalography (MEG) data (\eg~Plis \etal~\cite{Plis2011}) or metabolic brain networks from FDG-PET data (\eg\ Zhou \etal~\cite{Zhou2014}) were not included. Similarly, because the focus here is on either structural or functional connectivity, this review does not include studies which primarily examine relationships between cortical thicknesses in different regions (\eg\ Cuingnet \etal~\cite{Cuingnet2010}, Liu \etal~\cite{Liu2013}). Finally, papers that used machine learning to construct connectomes only and did not perform any further learning tasks on those connectomes (\eg\ Jiang \etal~\cite{Jiang2015}) were not included in Table~\ref{tab:depthReportTable}.

The remainder of this paper is organized as follows: We first breifly discuss how connectomes are constructed from fMRI or dMRI data (Section~\ref{sec:connectomeConstruction}) before presenting a thorough survey of different machine learning tasks that have been explored (Section~\ref{sec:tasks}), different machine learning models that have been proposed or employed (Section~\ref{sec:models}) and finally a summary of the current state of the art and a discussion about what we believe are strategic directions for future research (Section~\ref{sec:conclusions}).

\section{Connectome Construction}
\label{sec:connectomeConstruction}
Constructing a connectome from a dMRI or fMRI scan requires a pipeline of processing steps~\cite{Bullmore2009}. Here, we do not cover the MRI acquisition process, which has been discussed in great detail in previous works~\cite{mueller2015diffusion,Booth2015,VandenHeuvel2010}. Also, since the focus of this survey is on the methods that take connectomes as input, we only outline the most common methodologies for connectome construction and we don't discuss the details of each step. However, it should be noted that changing parameter settings or using different methods at each pipeline stage has been shown to have a significant impact on results~\cite{Yao2015,Liang2012,Zhong2015,Qi2015,Buchanan2014}. For in-depth explorations of how different construction methods affect resulting functional connectomes see Liang \etal~\cite{Liang2012} and for structural connectomes see Zhong \etal~\cite{Zhong2015}, Qi \etal~\cite{Qi2015} or Buchanan \etal~\cite{Buchanan2014}. Furthermore, Yao \etal\ examined specifically how atlas choice can affect resulting structural and functional connectomes~\cite{Yao2015}.

Formally, a connectome is a graph, $G(E,V)$, representing the structural or functional connectivity between pairs of brain regions of interest (ROI) or landmarks represented as a set of nodes, $V$. The connectivity between pairs of nodes is represented as the set of edges, $E$, which may be binary or weighted (and either directed or undirected). Each connectome is typically encoded as an adjacency matrix, $A \in \mathbb{R}^{|V| \times |V|}$, in which each entry represents either the existence of an edge (in a binary network) or the weight of an edge (in a weighted network). In certain cases, each node may also be assigned a weight (\eg\ Ng \etal~\cite{Ng2012} used fMRI signals to define the weight at each node and dMRI connecitivity to define the weight of edges between nodes).

The number of nodes, $|V|$, defines the scale of the brain network with large scale networks being defined by a few nodes from large ROIs (or spatially sparse landmarks) and fine scale networks being defined by many nodes from small ROIs (or densely distributed landmarks)~\cite{Alnæs2015}. In some extreme cases, each voxel in the image is associated with a node~\cite{Anderson2011}. Some works have also constructed connectome from multiple scales~\cite{Jin2015} or instead constructed a fine scale connectome and then employed multi-scale connectome features~\cite{Khazaee2015a,Khazaee2015}.

A common way to define ROIs is to register (\ie\ align, either linearly or non-linearly) each MRI brain scan to a template image with an associated atlas of segmented brain regions that can be used for parcellation. Many machine learning studies on both functional and structural connectomes used atlas based approaches to identify ROIs. In Table~\ref{tab:depthReportTable}, of the \nPapers~listed papers, \nPapersUsedAtlases~used atlases, such as the automated anatomical labeling (AAL) atlas~\cite{Tzourio-Mazoyer2002}, versus only \nPapersUsedLandmarks~that used landmarks, such as the dense individualized common connectivity-based cortical landmarks (DICCCOL)~\cite{Zhu2013}. The main advantage of using a labeled atlas of the brain to define ROIs is that it facilitates an objective anatomical interpretation of learned models which can easily be compared to other results in the literature. However, it has been noted in the literature that atlas based approaches are susceptable to imperfect segmentations of functional regions which can considerably impact sensitivity~\cite{Smith2011}. DICCCOL uses the white matter structure of the brain in attempt to ensure the correct placement of each landmark but only defines a single position, rather than a region, for every landmark which may limit the number of voxels that can be reliably associated with each connectome node. 

\nPapersUsedUnsupervised~papers, the majority of which focused on fMRI, used independent component analyis (ICA)~\cite{lee1998independent} or other unsupervised learning techniques such as Ward hierarchial clustering~\cite{ward1963hierarchical,Ng2012} in order to identify regions with self-similar features. The remaining papers either assigned nodes to hand-delineated ROIs, ROIs from uniform tile-like parcellations of the brain or to each voxel. One drawback of these latter approaches (including the learning approaches), is that the nodes are not \emph{apriori} associated with anatomically defined regions and so interpretation of results may be more difficult. Furthermore, when these kinds of approaches are applied independently to each scan, there may not be correspondence between ROIs in different scans~\cite{Takerkart2012}. This will also be the case when comparing connectomes constructed with different pipelines. However, there exist methods, such as graph kernels (Section~\ref{sec:graphKernels}), that can compare connectomes and be used to build learning models even when correspondence between nodes is not defined.

Edge definitions vary by modality so in the next two subsections we discuss how connectivity is respectively defined for dMRI and fMRI based connectomes.

\subsection{Structural Connectomes}
For structural connectomes derived from dMRI, edges between each pair of nodes are assigned (or are assumed to exist and are weighted) by measuring the degree of white matter connectivity between the associated pair of ROIs. This is typically done by first fitting a diffusion model, such as the diffusion tensor imaging (DTI) model to each voxel, and then reconstructing the white matter fibers via tractography~\cite{Booth2015}. The degree of connectivity between a pair of ROIs is then commonly defined as the number of reconstructed tracts with end-points in (or which simply pass through) both ROIs. In the case of proabilistic tractography, the degree of connectivity may represent the probability of a given tract connecting two ROIs~\cite{parker2003probabilistic,friman2006bayesian}. Alternatively, connectivity can be defined as the mean diffusion fractional anisotropy (FA) of each voxel intersecting the tract and then averaged over all tracts between the two ROIs~\cite{Mitra2016,Robinson2010}. Rather than choosing only one definition, some studies analyse features from multiple connectomes, constructed with different definitions of structural connectivity~\cite{Wee2011,Brown2015}. Finally, while tractography is the most popular way to define structural connectivity, other approaches have been explored including connectivity based on the fast marching algorithm over the set of voxels with diffusion defining edges between neighbouring voxels~\cite{Prasad2015,Li2011,booth2012multi}. For instance, Prasad \etal\ examined a measure of maximum flow between pairs of ROIs across a lattice defined by the diffusion tensor at each voxel~\cite{Prasad2015}. Because dMRI does not measure directionality of diffusion flow, edges of structural connectomes are necessarily undirected~\cite{Booth2015}. For further discussion on constructing structural connectomes, see Hagmann \etal~\cite{Hagmann2007} and de Reus and Heuvel~\cite{DeReus2013}. 

\subsection{Functional Connectomes}
Rather than measuring diffusion, fMRI data measures a blood-oxygen-level dependent (BOLD) signal that is indicative of neural activity over time, at each voxel. This signal is typically averaged within each ROI. Functional connectivity between a pair of ROIs is computed using some measure of correlation or dependence between the two averaged signals. Functional connectivity can then be interpreted as communication between pairs of brain regions. The most standard measure of correlation is the Pearson's correlation coefficient. However, in the case of $N \ll M$, Pearson's correlation overestimates the number of correlated pairs of regions~\cite{Yoldemir2015coupled}, especially when the data is noisy as is the case with fMRI ~\cite{Sweet2013}. Gellerup \etal\ performed a comparison of twelve different measures of connectivity and found Pearson's correlation with Bonferroni multiple comparison correction, which reduces the number of false positives by imposing a very strict significance threshold, worked best~\cite{Gellerup2015}. Instead, some studies defined the weight of each edge as the partial correlation between two ROIs. To compute the partial correlation between two BOLD time signals, the effect of every other time signal is first removed (via regression) and then Pearson's correlation is computed. Rosa \etal\ showed that using partial correlation based functional connectomes lead to higher prediction accuracy~\cite{Rosa2013}. Another approach is to estimate a sparse inverse covariance matrix for the set of ROI time signals. One commonly used method of this type is called the graphical least absolute shrinkage and selection operator (GLASSO) which uses $\ell_1$ regularization to promote a sparse solution~\cite{Friedman2008}. Smith \etal\ and Rosa \etal\ found that functional connectome construction methods based on sparse estimation of covariance work best in terms of accurate estimation of true connections and in terms of classification accuracy, respectively~\cite{Smith2011,Rosa2013}. 

After computing the correlations between ROIs, many methods (\eg~\cite{Jie2014,Fei2014}) apply Fischer's r-to-z transform~\cite{Silver1987} which converts correlation values into Z-scores in order to prevent bias from being introduced in subsequent steps. Furthermore, since correlations (and Z-scores) can be negative but certain feature descriptors may assume non-negative inputs, some approaches applied an absolute value function to correlation values, but extracted features from edges with positive and negative correlations separately~\cite{Anderson2014}.

Edges in functional connectomes are typically undirected as they are in structural connectoms. However, because fMRI has a temporal component, it is possible (but less common) to construct functional connectomes with directed edges~\cite{Shahnazian2012,Wee2013a}. These connectomes represent what is known as effective connectivity. Both Shahnazian \etal~\cite{Shahnazian2012} and Wee \etal~\cite{Wee2013a} trained learning models on directed, Granger causality networks~\cite{Granger1988}. To infer this causality, Shahnazian \etal\ used conditional Granger causality analysis (CGCA) whereas Wee \etal\ used multivariate autoregressive (MAR) modelling\cite{Shahnazian2012,Wee2013a}. Smith \etal\ found that using Granger causality and other temporal lag based methods performed poorly when trying to infer directional structure, whereas Patel's $\tau$ measure~\cite{Patel2006} (which examines the imbalance between conditional probablities of each direction of connectivity), performed best~\cite{Smith2011}. Effective connectivity networks better model the true dynamics of the brain~\cite{Friston2009} and have been shown to enable higher classification accuracy~\cite{Wee2013a}, but come with the drawback of requiring twice as many edge weights to represent them.

\subsection{Signal Noise and Bias}
There exist many sources of noise and bias in the construction process for both structural and functional connectomes. Inaccurate placement of ROIs~\cite{Smith2011} and patient motion during scan, due to the long scan times required for both dMRI and fMRI~\cite{Rohde2004,VanDijk2012,Power2012}, can affect connectomes of both modalities. Though, whereas structural connectomes tend to contain more false negative connections, due to tractography algorithms terminating too early in crossing and heavily curved white-matter regions, functional connectomes tend to contain more false positive connections, in part due to patient breathing and pulse (which affect blood oxygen levels across the brain)~\cite{Yoldemir2015coupled}.

\section{Machine Learning Tasks}
\label{sec:tasks}
The majority of studies that apply machine learning to connectome data have focused on predicting outcomes via classification or regression (\nPredictionPapers~of \nPapers~of the papers in Table~\ref{tab:depthReportTable}). However, other works have explored applying machine learning to unsupervised clustering of patient connectomes into groups, identification of important subnetworks and optimization of ROIs.

\begin{figure*}
\centerline{\includegraphics[width=0.95\textwidth]{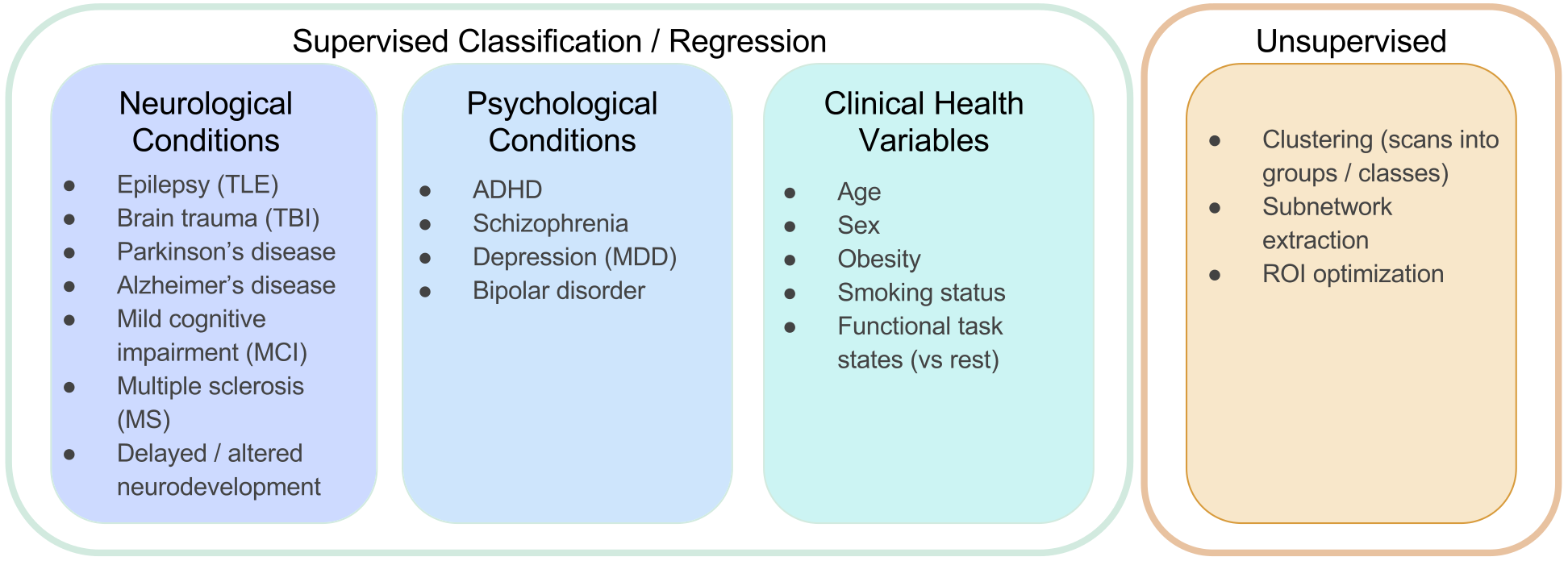}}
\caption{Machine learning tasks that have been performed using MRI based connectome data.}
\label{fig:tasks}
\end{figure*}

\subsection{Outcome Prediction}
\label{sec:outcomePrediction}
Table~\ref{tab:depthReportTable} reveals that features of structural and functional connectome data can be used to identify or predict a wide variety of pyschological disorders, neurological disorders, patient attributes and mental states (see Fig.~\ref{fig:tasks}). For instance, connectome features have been used to identify patients with schizophrenia~\cite{Dodero2015,Shen2010,Arbabshirani2013,Bassett2012,Zhu2014}, major depressive disorder (MDD)~\cite{Sacchet2015,Rosa2013,Rosa2015,Craddock2009,Guo2012}, attention deficit hyperactivity disorder (ADHD)~\cite{Cheng2012a} and bipolar disorder~\cite{Moyer2015}. Clinical diagnoses of these psychological disorders can be challenging and accurate connectome based identification could allow earlier diagnosis and treatment~\cite{Arbabshirani2013}. While the majority of these works used functional connectomes to train their models, structural connectomes were used by Sacchet \etal~\cite{Sacchet2015} and Moyer \etal~\cite{Moyer2015} for identifying MDD and Bipolar disorder, resepectively.

The use of machine learning on connectome data for early diagnosis could especially be beneficial for neurodevelopmental disorders such as austism spectrum disorder (ASD)~\cite{Li2012,Jin2015,Wee2016} and delayed motor and cognitive development in preterm infants~\cite{Brown2015,Brown2016,kawahara2016} since early intervention of these conditions can improve patient outcomes~\cite{bear2004early,Wee2016}. The use of connectome data and machine learning models for identification of neurodegenerative diseases such as Parkinson's disease~\cite{Galvis2016,Gellerup2015}, Alzheimer's disease (AD)~\cite{Prasad2014,Prasad2015,Zhan2015,Moyer2015,Khazaee2015a,Khazaee2015,Wee2013a,vanderweyen2013,chen2011,Ashikh2015,Yang2015} and early and late mild cognitive impariment (MCI)~\cite{Fei2014,Zhu2014,Jie2014,Jie2014a,Jie2014b,Wee2011,Wee2012,Wee2013,Zhu2014a} may also be important as differential diagnoses of these conditions using traditional image analysis techniques can be challenging~\cite{reitz2014alzheimer}. Furthermore, the trained machine learning models may reveal important (\ie\ discriminative) connections and subnetworks in the brain which could help researchers better understand disease etiology.

Connectome features have also been used to train models that can discriminate normal control (NC) infants from preterm born infants~\cite{Ball2016} and infants with neonatal encephalopathy~\cite{Ziv2013} as well as adult NCs from multiple sclerosis (MS) patients~\cite{Richiardi2012}, patients who have had traumatic brain injury (TBI)~\cite{Mitra2016,vanderweyen2013,Caeyenberghs2013} and patients with temporal lobe epilepsy (TLE)~\cite{Munsell2015,Kamiya2015}. As well, learned models have been used to predict the location of epileptic regions~\cite{Sweet2013}. Machine learning models trained on connectome features can also accurately identify patient age~\cite{kawahara2016,Qiu2015,Robinson2010,Pruett2015,Smyser2016,Dosenbach2011,Ghanbari2014}, sex~\cite{Solmaz2016}, obesity~\cite{Park2015}, status as a smoker or non-smoker~\cite{Pariyadath2014} and restedness~\cite{Kaufmann2016}. Age, in particular, is an interesting application since a person's physiological or developmental age may differ from their actual age~\cite{mitnitski2002frailty}. A discrepancy between predicted and true age may indicate delayed or accellerated brain development or aging.

Finally, functional connectome data, in particular, has been used to train models to distinguish between different functional task loads (including resting)~\cite{Mokhtari2012,Shahnazian2012,Alnæs2015}, between rest and stimulated states (\eg\ watching a movie~\cite{Richiardi2011}, viewing a face~\cite{Ng2012}, hearing auditory stimuli~\cite{Vega-Pons2014,Takerkart2012}) and before, during and after memory tasks~\cite{Ng2016}.

Table~\ref{tab:depthReportTable} shows that the majority of studies in this field have focused on adult and elderly age groups. This is reflected in the large number of studies on conditions like AD and MCI, which primarily affect aging populations. Fig.~\ref{fig:agesVsYear} shows, however, that the number of studies focused on infant, child and adolesent age groups is increasing.

Of the \nPredictionPapers~studies that performed prediction, \nClassificationPapers~performed classification and only \nRegressionPapers~performed regression. Of those \nClassificationPapers~classification studies, \nTwoClassPapers~performed binary (2-class) classification and the remaining \nMultiClassPapers~studies performed multi-class classification. Different tasks require different machine learning models, which we discuss below in Section~\ref{sec:learningModels}.\\

\begin{minipage}{.48\textwidth }
\includegraphics[angle=0,width=0.99\textwidth]{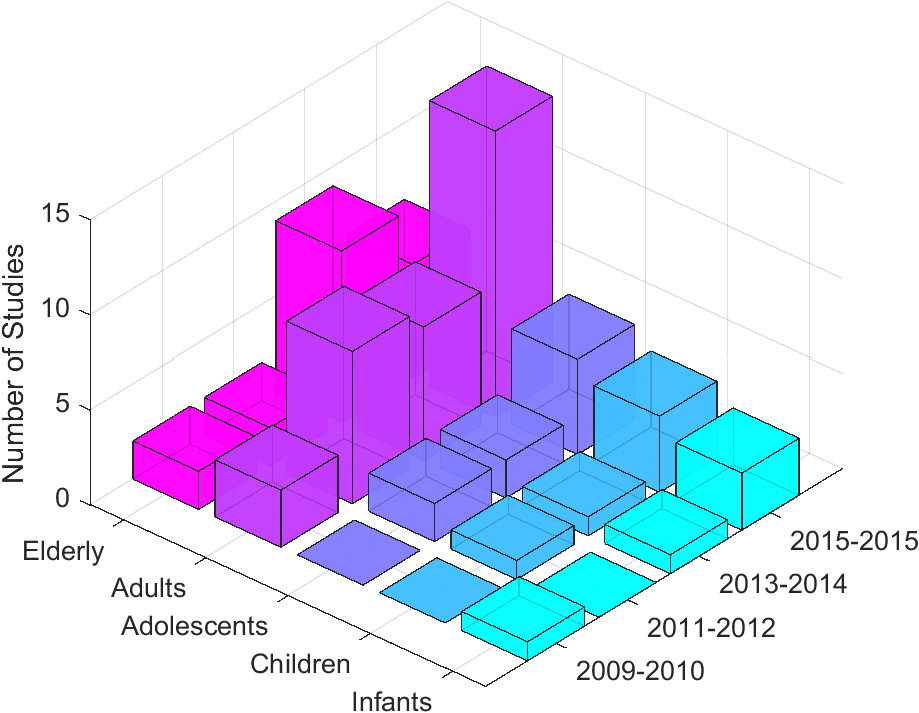}
\captionof{figure}{2D histogram of number of studies grouped by age group of study and biennially by year of publication. Note that there are an increasing number of studies not only on adults but also on adolescents, children and infants.}
\label{fig:agesVsYear}
\end{minipage}
\begin{minipage}{.02\textwidth }
~
\end{minipage}
\begin{minipage}{.48\textwidth }
\includegraphics[angle=0,width=0.99\textwidth]{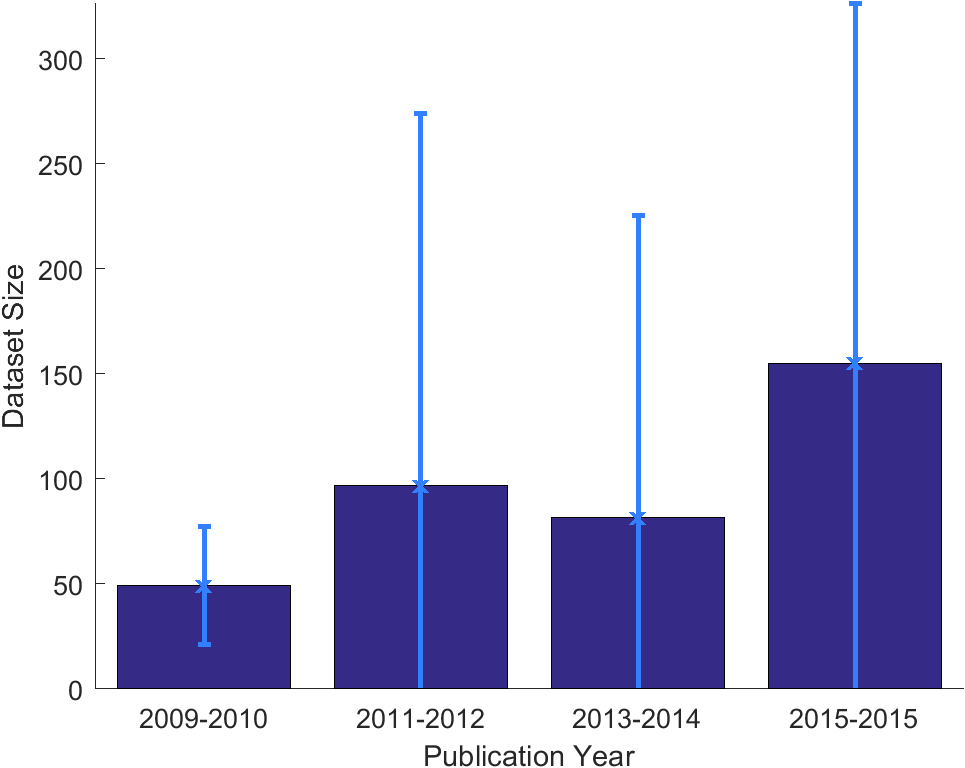}
\captionof{figure}{Average number of MRI scans per study versus publication year, grouped biennially.}
\label{fig:numScansVsYear}
\end{minipage}\\

As was mentioned above, most studies that have applied machine learning models to connectome data have been exploratory (rather than large scale studies), and have trained and validated their models on small datasets only. However, Fig.~\ref{fig:numScansVsYear} shows that the mean number of scans used per study is growing, as is the standard deviation. This may be in part due to the recent emergence of large public databases such as the human connectome project (HCP)~\cite{VanEssen2013}, the alzheimer's disease neuroimaging initiative 2 (ADNI-2)\footnote{\url{http://www.adni-info.org/}}(which only began collecting dMRI and fMRI data in 2011), the autism brain imaging data exchange (ABIDE)~\cite{di2014autism} and the ADHD-200 dataset\footnote{\url{http://fcon_1000. projects.nitrc.org/indi/adhd200/}}. Pre-processed connectome data is available for both of the latter two datasets from the pre-processed connectome project\footnote{\url{http://preprocessed-connectomes-project.org/}}, also initiated in 2011.

The average best reported prediction accuracy across all datasets in all classification studies was $81\%$. Note however that the difficulty of the prediction task and a variety of other factors can greatly impact prediction accuracy. For instance, Iidaka found that the number of folds in the cross-validation scheme can significantly affect the prediction accuracy (Section~\ref{sec:validation})~\cite{Iidaka2015}. We found that prediction accuracies were moderately but significantly negatively correlated with publication year, $r=-0.419, p=8\times10^{-5}$. This may be partially due to the effect of increasing dataset sizes, as the prediction accuracies are weakly but significantly correlated with the number of scans, $r=-0.291, p=0.007$. However, it is likely also due to researchers exploring increasingly challenging problems such as multi-class classification (\eg\ Prasad \etal~\cite{Prasad2015}, Takerkart \etal~\cite{Takerkart2012}) and prediction of future outcomes (\ie\ cases where there is a temporal delay between acquiring the scan and labelling the scan)~\cite{Brown2015,Brown2016,kawahara2016}.

\begin{figure*}
\centerline{\includegraphics[angle=0,width=0.80\textwidth]{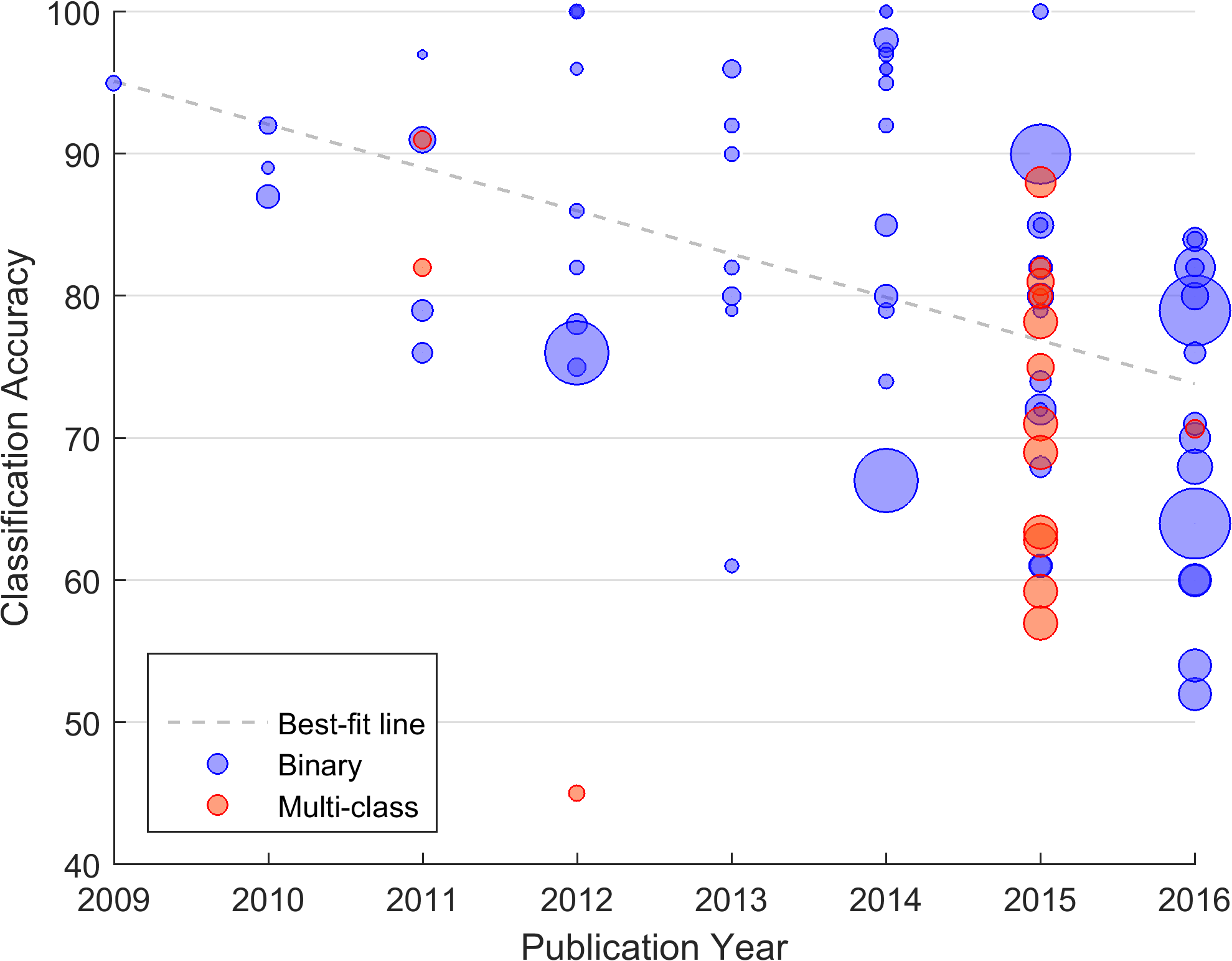}}
\caption{Best reported accuracy of each dataset in each classification study vs year of publication. Size of marker represents the relative size of the dataset. Binary and multi-class classification tasks are denoted by blue and red markers respectively.}
\label{fig:accVsYear}
\end{figure*}

\subsection{Clustering, Subnetwork Extraction and ROI Optimization}
A minority of machine learning papers employing connectome data did not perform outcome prediction. In place of supervised learning models, Ashikh \etal~\cite{Ashikh2015} used an unsupervised approach to group connectomes into clusters and found that these clusters matched the class labels (NC, early MCI, late MCI and AD). Other studies used connectome data with unsupervised techniques to learn subnetworks containing regions that co-varied across subjects~\cite{Yoldemir2015} and subgroups of subjects~\cite{Gao2015} or to learn subnetworks discriminative of TBI~\cite{Chen2015a}, post-traumatic stress disorder (PTSD)~\cite{Li2014b}, ADHD~\cite{Anderson2014} or ASD~\cite{Ghanbari2014}. In the work by Ghanbari \etal\, different subnetworks were found that were predictive of age and ASD, simultanenously~\cite{Ghanbari2014}. Yang \etal's method finds subnetworks that are common across the entire dataset but identifies those connections that are specific to individual groups~\cite{Yang2015}. Chen \etal\ used connectome based models to optimize the ROIs for group-wise consistency of connectivity patterns~\cite{Chen2014a}.

\section{Machine Learning Models for Connectome Data}
\label{sec:models}
A wide variety of machine learning methods have been explored for use on structural and functional connectome data. Given the unique characteristics of network type data (discussed in Section~\ref{sec:challengesOpportunities}), many approaches present ways to both reduce dimensionality and leverage the data's special topology. In this section we discuss methods to deal with class imbalance, connectome features, methods of feature selection and dimensionality reduction, learning models, kernels, regularization terms and validation procedures that have been proposed for use on connectome data.

\subsection{Class Imbalance}
\label{sec:classImbalance}
When training a learning model, it is often ideal to have a balanced number of examples in each class (or a balanced histogram of labels in the case of continuous labels) in order to not introduce bias into the model. However, of the papers in Table~\ref{tab:depthReportTable}, \nImbalancedPapers~validated their approach on datasets where the largest class was more than twice as large as the smallest class (which can be considered very imbalanced). Some of these studies employed strategies to mitigate the effects of class imbalance. For instance, Ball \etal\ used only a subset of the scans in their majority class, along with all of the scans in their minority class when using ICA to determine ROIs for connectome construction~\cite{Ball2016}. In contrast, Kaufmann \etal\ noted that despite their imbalanced classes, they performed ICA on the whole dataset, preferring to include all scans in the definition of ROIs~\cite{Kaufmann2016}. (Neither study validated their choice experimentally.)  

When training their prediction model, Ball \etal\ also weighted training instances from the minority class proportionally higher, which is similar to replicating minority class instances until the number of instances in each class is equal. Instead, Khazaee \etal\ used the hold-out method~\cite{fielding1997review}, selecting a balanced, random subset of training data in each round of cross validation~\cite{Khazaee2015}.

Kawahara and Brown \etal\ used the synthetic minority oversampling technique (SMOTE)~\cite{chawla2002smote} to generate synthetic instances in their training set~\cite{kawahara2016,Brown2016}. They generated more synthetic instances of the minority class than those in the majority class in order to balance their training set before training their classifier. Similarly, Brown \etal\ used their proposed local synthetic instances (LSI) to augment their dataset, generating only instances of the minority class~\cite{Brown2015}. LSI assumes the case of HDSSS. In contrast to SMOTE, which attempts to sample from a manifold of training data that may be too sparsely defined, LSI generates new synthetic instances which are guaranteed to be near to real instances and that are perturbed in a realistic way~\cite{Brown2015}.

Class imbalance can also affect validation as well, biasing traditional measures such as prediction accuracy, sensitivity and specificity. Jie \etal, Wee \etal, Galvais \etal\ and Ball \etal\ all validated their models using balanced accuracy, defined as the mean of the sensitivity and specificity, which weights the accuracy in each class equally giving a better sense of how the prediction model actually performed on imbalanced test data~\cite{Jie2014,Jie2014b,Wee2012,Galvis2016,Ball2016}. Brown \etal\ instead selected only balanced test sets for each round of validation~\cite{Brown2015,Brown2016}.

Class imbalance is not a unique problem to machine learning on connectome data, but the problem is exacerbated when combined with the HDSSS problem, which is common for connectome datasets. More research is required to find new methods (such as LSI) that can balance and augment connectome training sets in a principled and effective way.

\subsection{Connectome Features}
\label{sec:connectomeFeatures}
In order to train a model, useful features are first extracted from the connectome (or the connectome as a whole can be used, given a proper representation). Similar to features from grid-like image data, connectome features may range in scale and complexity.

At the finest scale, the feature set may comprise a vector embedding of the edge-wise features in the connectome adjacency matrix. Of the papers in Table~\ref{tab:depthReportTable}, nearly half (\nEdgeFeaturePapers~of \nPapers) trained machine learning models on basic edge-wise features of either structural connectivity (\eg\ ROI-pair tract count or mean FA) or functional connectivity (\eg\ correlation between mean fMRI signals in pairs of ROIs). Park \etal\ employed a multi-modal approach and defined connectivity between two ROIs in terms of both structural fiber density and mean nodal functional connectivity of the two ROIs~\cite{Park2015}. Similarly, Chen \etal\ constructed both structural and functional connectomes for each subject and used both connectome types as input to their model~\cite{Chen2015a}. By passing raw edge features into the learning model, no \emph{a priori} assumptions need to be made about which aspects of the connectome are important. This puts more responsibility on the learning model to extract relevant cues from the data. Additionally, each individual edge feature may be noisy ~\cite{Sweet2013,Richiardi2013} and since the number of edges is often large (\ie\ in this case $M = \mathcal{O}(N^2)$), feature selection becomes especially important.

One way of reducing the effect of noise and uncertainty on connectivity measurments is to threshold the value of each edge at an assumed or calculated noise amplitude, resulting in a binary value for each edge weight~\cite{Vega-Pons2014, Wang2014, Jie2014b}. Extending this idea, Jie \etal\ used features of binary connectomes generated from multiple thresholds~\cite{Jie2014}.

Some works on fMRI used edge features but also introduced temporal information. Wee \etal\ used edge features from regional fMRI signal correlations computed in a set of overlapping temporal windows~\cite{Wee2013}. They used a fused multiple GLASSO (FMGL) approach to jointly estimate the signal covariance matrices in all temporal windows simultaneously and edge features from each window were concatened into one large feature vector. In their recent follow-up~\cite{Wee2016}, rather than using a sliding window, they first clustered the fMRI signals temporally in order to find distinct brain states before estimating the functional connectivity networks. Ashikh \etal\ used a sliding temporal window but extracted the temporal variance in functional connectivity at each edge as the input features to their learning model~\cite{Ashikh2015}. While these approaches can better capture dynamic patterns in brain activity, they require estimation of functional connectivitiy from fewer time slices, since a window or temporal cluster is a strict subset of the entire set of time slices, and so may fit the data less well.

Other works in the literature have proposed using features of each node, including features of local connectivity~\cite{Guo2012,Kamiya2015,Latora2001,Wang2014,Jie2014,Jie2014b,Wee2011,Anderson2014,Li2014b}. For instance, from each node in each connectome, Guo \etal\ and Kamiya \etal\ extracted nodal degree, local efficiency and betweenness centrality~\cite{Guo2012,Kamiya2015}. Given a node, $i$, and its set of neighbouring nodes, $\mathcal{N}(i)$, the (weighted) nodal degree at node $i$, is defined as,
\begin{equation}
d_i = \sum_{j \in \mathcal{N}(i)}{w_{ij}}
\end{equation}
\noindent where $w_{ij}$ is the weight of the edge between nodes $i$ and $j$. Nodal degree captures how well connected a node is to the rest of the connectome. Similarly to edge-wise features, nodal degree features have also been computed within a sliding window across time~\cite{Li2014b}. Local efficiency (LE) involves the concept of a shortest path, along edges and is defined as,
\begin{align}
LE_i = \frac{1}{N}\sum_{j,k \in \mathcal{N}(i) }{\frac{\left(w_{ij} w_{ik}  |P^*_{j,k}(\mathcal{N}(i))|^{-1} \right) ^{\frac{1}{3}}}{d_i (d_i - 1)}  }
\end{align}

\noindent and where $P^*_{j,k}(\mathcal{N}(i))$ is the shortest path of edges, $(p,q) \in E$, between nodes $j$ and $k$, passing only through nodes neighbouring $i$~\cite{Rubinov2010,Latora2001}. $|P_{j,k}^*(\mathcal{N}(i))|$ is then the length of this shortest  path, computed as,

\begin{equation}
|P_{j,k}^*(\mathcal{N}(i))| = \sum_{(p,q) \in P^*_{j,k}(\mathcal{N}(i))}{\frac{1}{w_{p,q}}}.
\end{equation}

\noindent Intuitively, local efficiency measures how robust a node's local neighbourhood is. Given the set of shortest paths between every pair of nodes in the entire connectome, betweenness centrality at node $i$ is the fraction of those paths that pass through node $i$~\cite{Rubinov2010}. Another local measure of network topology, called the clustering coefficient (CC), is defined in terms of triangles of edges between a node, $i$, and two of its neighbours, $j$ and $k$. It is computed as,
\begin{align}
CC_i = \frac{2 t_i}{d_i (d_i - 1)} && \text{where} && t_i = \frac{1}{2}\sum_{j,k \in\mathcal{N}(i)} (w_{ij} w_{jk} w_{ik})^\frac{1}{3}.
\end{align}
\noindent Note that $t_i$ can be interpreted as the average geometric mean of weights of edges forming triangles around $i$. If the edge weights are binary then the same formula holds and $t_i$ counts the number of triangles around $i$. CC has been used as a node-wise feature by several studies~\cite{Kamiya2015,Wang2014,Jie2014,Jie2014b,Wee2011} and indicates the degree of segregation that a node and its direct neighbours have from the rest of the newtork. Hi values of CC indicates that a node is part of a segregated module. In contrast, the participation coefficient measures the degree of integration a node has with such segregated modules and so is a measure of centrality, similar to betweenness centrality~\cite{Rubinov2010}. Nodes with high participation coefficients are then likely to be interconnecting hubs. Anderson \etal\ extracted the participation coefficient at each node, along with non-connectome features of cortical thickness and non-image-based metadata~\cite{Anderson2014}. Lastly, some fMRI studies simply used the BOLD signal at each node as features~\cite{Yoldemir2015,Yoldemir2015coupled}. In these cases, the connectome nature of the data was not explicitly encoded in the input features, but instead it was encoded into the learning models as prior information. Note that if a constant number of features is extracted at each node, there will be $M= \mathcal{O}(N)$ features versus $M= \mathcal{O}(N^2)$ edge-wise features.

Somewhat similar to measures of nodal CC and participation coeficient, Tunc \etal\ used features of connectivity within and between pre-defined functional subnetworks~\cite{Solmaz2016}. In contrast to nodal CC and participation coeficient features, these features were only extracted for each single and pair of pre-defined subnetworks, rather than for each node in the entire connectome.

Rather than using pre-defined subnetworks, Fei \etal\ extracted the most frequently occuring subnetworks in each class using the graph-based substructure pattern mining (gSpan) algorithm~\cite{Fei2014}. These subnetworks were then extracted across all connectomes of both classes. While these subnetworks are not necessarily discriminative, they each represent stable parts of the network structure of at least one class. From this set of features, only the most discriminative were selected (Section~\ref{sec:featureSelection} below).

At the largest scale, single features of the entire connectome can be extracted (\ie\ $M = \mathcal{O}(1)$)~\cite{Ziv2013,Prasad2015,Sacchet2015,Brown2014,Chung2016}. Global network measures are often based on local (nodal) network measures, but instead summarize the entire network toplogy and include measures of network integration, segregation, small-worldness and rich-clubness~\cite{Rubinov2010,van2011rich}. Brown \etal, Prasad \etal, Sacchet \etal\ and Ziv \etal\ all extracted multiple global network measures as input features to learning models~\cite{Brown2014,Prasad2015,Sacchet2015,Ziv2013}. In addition, Ziv \etal\ also extracted a different type of global feature of the network based on subgraph enumeration~\cite{Ziv2013}. To compute this feature, they count the number of possible subgraphs that can be generated by a walk of length 8 (\ie\ 8 edges) on a connectome with thresholded, binary edge weights. These counts are binned by isomorphic subgraphs, and the array of bin counts is used as a feature vector.

The concept of heat flow has also been used to extract global, topological connectome features~\cite{Chung2016}. Chung \etal\ simulated heat flow across the network domain of the connectome and measured the maximum heat flow rate, the time to achieve maximum heat flow and the time to achieve near equilibrium.

While these global measures capture salient features of the entire connectome topology in a very compact form and have been successfully used for machine learning tasks on connectome data, they may be less sensitive to local cues and so may not be suitable for applications where local effects are important. Furthermore, while each global measure may have a well undestood topological interpretation, these functions are typically non-invertable. Thus, for machine learning models trained on global network features, analysis of the influence of individual edges may not be possible, making anatomical interpretation and localization of important regions more difficult.

Rather than reducing dimensionality with global measures, Ng \etal\ treated each adjacency matrix, $A$, (computed from estimation of functional covariance) as a single multi-variate feature representing a point on the manifold of positive definite matrices~\cite{Ng2016}. Dodero \etal\ applied the same approach but first computed the graph Laplacian, $L = D - A$, of each connectome, where $D$ is a diagonal matrix with the degree, $d_i$ of each node, $i$, along the diagonal~\cite{Dodero2015}. In contrast to approaches that collapse edge features into a single vector, these methods retain the matrix structure of the original connectome representation enabling them to use the manifold in which they are embedded.

Finally, while all of the above features have been engineered, high level features may also be learned by a feature selection or dimensionality reduction technique or by the learning model itself. Engineered features can be understood as methods for dimensionality reduction or low level feature selection that incorporate some prior knowledge about the expected structure of the data. However, prior knowledge can be incorporated in other ways, such as model regularization. Throughout the next few sections, different methods for learning high level connectome features are discussed, of which many require basic (low level) edge features as input.

\subsection{Feature Selection and Dimensionality Reduction}
\label{sec:featureSelection}
In order to train a model that fits well to the data and also generalizes well to unseen instances, it is useful to select or combine features in a way that retains only the most informative (\eg\ discriminative) and least noisy features. While a connectome representation of the brain is relatively compact compared to, for instance, a volume of voxels (\eg\ typically $10^3$ to $10^5$ connections in a connectome vs $10^6$ to $10^7$ voxels in an image volume), feature selection and/or dimensionality reduction remain important aspects of a machine learning approach, especially when many low-level features are being used and when the number of datasets to train on are limited. Feature selection methods can broadly be split into three categories: Filters, wrappers and embedded selectors (summarized next and examined in detail in Sections~\ref{sec:filters},~\ref{sec:wrappers} and~\ref{sec:learningModels}). 

Wrapper type feature selectors (Section~\ref{sec:wrappers}) are based on a learned model (often a classifier) that is fit to some subset of the training data and identifies discriminative features as those that were most important (\eg\ for classification). While these methods can be slower, because they require a model to be trained, often multiple times, they have the advantage of identifying interactions between features that might be independently weak. In contrast, filter type feature selectors (Section~\ref{sec:filters}) do not train a model and typically decide the importance or weight of each input feature based on a heuristic criterion (often independently)~\cite{brown2012conditional}. These types of feature selectors are then efficient and can often be run in parallel across the data, but may discard features that are discriminative when combined together. Embedded feature selectors are similar to wrappers except that they are not run as a preprocessing step but instead are integrated into the final learning model being trained, usually as a regularization term that encourages only a sparse subset of the features to be used. We discuss embedded feature selectors in terms of how they regularize different learning models in Section~\ref{sec:learningModels}. Some approaches used multiple feature selection techniques, stacked in order to utilize the advantages of each technique~\cite{Park2015,Zhu2014,Khazaee2015}.

Along with feature selection, a variety of dimensionality reduction techniques (discussed below in Section~\ref{sec:dimensionalityReduction}) have been applied to connectome data in preparation to machine learning tasks. While similar to feature selection in that the number of output features is reduced, dimensionality reduction techniques act as transforms on the input feature space, rather than filters of individual features.

\subsubsection{Filter Type Feature Selectors}
\label{sec:filters}
Of the papers listed in Table~\ref{tab:depthReportTable}, the most widely used feature selection technique is the application of a non-paired two-sample t-test to each edge feature~\cite{Zhu2014,Jie2014,Park2015,Richiardi2011,Mitra2016,Dosenbach2011}. A given edge feature is then only selected if the distributions of feature values (\ie\ often edge connectivity strength) across connectomes in the two classes are significantly different, indicating that the feature discriminates between classes. The use of this test for brain networks has also been called a brain-wide association study (BWAS) after the similar genome-wide association study (GWAS), commonly used in genomics~\cite{Ji2011}. The two-sample t-test, however, assumes that the number of instances in each class (\ie\ sample sizes) are equal, which is not true in general. Khazaee instead used the Fisher score~\cite{duda2012pattern}, which takes the size of each class into account~\cite{Khazaee2015}. 

The t-test also assumes normally distributed data which may not be true in practice, especially for features derived from data with a lower and/or upper bound (\eg\ edge tract counts, FA and correlation values). In order to remove the assumption of normally distributed data, Chen \etal~\cite{chen2011} used the non-parametric Wilcoxon rank sum test~\cite{noether1991wilcoxon}. Shen \etal\ used another non-parametric test, the Kendel tau rank correlation coefficient (KTRCC)~\cite{kendall1948rank} which, for each feature, counts the number of samples that co-vary with their label versus the number of samples that do not~\cite{Shen2010}. Features are then ranked by this measure and only the top ranking features are selected. Similarly, Park \etal~\cite{Park2015} employed permutation testing~\cite{Smith2013} which is also non-parametric but is often computationally intensive. These kinds of tests are adventageous because no assumption needs to be made about the distribution of edge weights across scans.

To achieve a given statistical significance level, it is assumed that only a single test has been run, which is not the case for tests like the t-test or Fisher score which are run independently for each feature (\ie\ M tests). False discovery rate (FDR)~\cite{Benjamini2005} has been used with the t-test as a correction for multiple comparisons, in order to account for the number of edges which may be selected by random chance and may not be class discriminative~\cite{Richiardi2011}. Multiple comparison corrections allows the significance threshold for feature selection to be picked in a more principled way.

However, all of the filter type feature selection tests discussed above, except permutation testing, assume that each feature is distributed independently of other features. This assumption is unlikely to hold for edge features, especially between those edges that share a node~\cite{Zalesky2010}. Correlation-based feature selection (CFS)~\cite{Hall1999}, used by Zhu \etal, selects those features that are not only group discriminative, but that are also more \emph{weakly} correlated with other features~\cite{Zhu2014}. This strategy helps reduce the number of redundant features that are selected. Similarly, Gellerup \etal\ employed minimum redundancy maximum relevance (mRMR) feature selection~\cite{Peng2005} that uses an objective function to select a set of features with maximum mutual information between features and labels and minimum mutual information between pairs of features~\cite{Gellerup2015}. mRMR is different from the rest of the above filter methods in that it examines all of the fearures simultaneously in order to choose which features to remove. Lastly, Mitra \etal\ used a t-test with network-based statistics (NBS)~\cite{Zalesky2010} which removes the assumption of indpendence between edge features by taking the network topology of the feature domain into account~\cite{Mitra2016}.

While a wide range of filter type feature selection methods have been used for machine learning on connectome data, there exist many other methods (\eg\ the Chi-squared test of independence) which have not been used by the papers in this review. Furthermore, of the methods that have been used, it is not clear which one is best. Nevertheless, based on the best attributes of the methods discussed above, it seems that an ideal filter type feature selector for connectome data should 1) select the most discriminative features, 2) reject redundant features, 3) account for dependencies between features imbued by the network structure of the data, 4) account for multiple comparisons and 5) not assume any specific distribution of feature values.

\subsubsection{Wrapper Type Feature Selectors}
\label{sec:wrappers}
One of the main drawbacks of filter type feature selectors is that the selected features are not validated for their combined discriminative power as part of the selection process. In contrast, wrapper type feature selectors simultaneously select and validate the set of selected features by computing some measure of fitness for the entire set, often in the form of a classification error. In this section, we first discuss wrapper based selectors that perform this validation using the accuracy or error of a prediction model that is trained over multiple subsets of the full set of features. Next we discuss wrapper based feature selectors that train a prediction model but use only the weights of that model (possibly summed over multiple rounds) to decide which features to keep. Finally, we discuss a wrapper based feature selector that does not train a prediction model but instead uses an objective function that is designed specifically to retain the best features.

To begin, a variety of papers have employed the popular recursive feature elimination (RFE) method~\cite{Guyon2002} which, using some subset of the training data, recursively trains classifiers on features sets, each with one feature removed~\cite{Wee2011,Li2012,Wee2013a,Ziv2013,Wang2014,Jie2014b,Jie2014,Pariyadath2014}. RFE then chooses the subset of features for which the classification error changed the least (up or down). This process repeats until only a specified number of features remain. Typically, the classifier used by RFE is the same type as the one that will be finally trained.

Fei \etal\, instead used the gSpan algorithm to generate a set of candidate subnetworks and then used discriminative subnetwork mining (DSM) to select only those subnetworks which led to the highest classification accuracy when passed to a graph kernel based classifier (see Section~\ref{sec:graphKernels}, below)~\cite{Fei2014}.

DSM and RFE are part of a group of methods that is called backwards sequential feature selection because the algorithms begin with a full set of features and progressively remove them. The forward sequential feature selection (FSFS) approach, used by Khazaee \etal\, begins instead with an empty set of features and iteratively adds features that produce the largest improvement in the fitness function (which in their case was defined by classification accuracy on a subset of the training data)~\cite{Khazaee2015}.

There is another group of wrapper type feature selectors that don't explicitly use the fitness function of the model but instead used the trained weights of a model (\eg\ classifier) to decide the importance of each feature. For instance, recursive feature ranking, used by Mokhtari \etal, is similar to RFE in that it recursively removes features by training a model~\cite{Mokhtari2012}. However, instead of using the classification error to determine the importance of each feature, at each round, the recursive feature ranking examines the weight placed on each feature by the trained classifer, and removes the feature with the smallest weight magnitude. Note, that for linear models, the weight with the smallest magnitude will also be the weight that changes the cost function the least when set to zero, and so these two feature removal criteria will be equivalent.

Vanderweyen \etal\ and Wee \etal\ both used a sparse linear regressor called the least absolute shrinkage and selection operator (LASSO) which uses an $\ell_1$ regularization term (described in Section~\ref{sec:learningModels}, below) to predict the class labels~\cite{vanderweyen2013,Wee2016}. After training on a subset of the training data, only those features with non-zero model weights are retained for the final prediction model. The ElasticNet method~\cite{Zou2005a}, used by Munsell \etal, uses the same sparse linear regressor as LASSO but adds an $\ell_2$ regularization term called a smoothness term because it discourages the weight on any one feature from becoming very large~\cite{Munsell2015}. In addition to ElasticNet, Munsell \etal\ also tried using sparse cannonical correlation analysis (SCCA)~\cite{hardoon2011sparse} and a deep auto-encoder based feature selector~\cite{Hinton2006} but found that ElasticNet outperformed both methods on their dataset~\cite{Munsell2015}. Note that BrainNetCNN, proposed by Kawahara and Brown \etal~\cite{kawahara2016}, also uses a deep neural network for feature selection but the feature selector is embedded into the learning model, so we discuss this model below in Section~\ref{sec:learningModels}.

Kamiya \etal\ used a feature selector called the Dantzig selector~\cite{Candes2007} that also involves an $\ell_1$ term over the model weights, but that is subject to a constraint that requires the maximum prediction residual, scaled by feature values for that connectome, to be less than a constant~\cite{Kamiya2015}. Those features with non-zero model weights are then passed to the final classifier. The Dantzig selector was designed specifically for cases where the number of features, $M$, is much larger than the number of scans, $N$, which is often the case in connectome studies. The manifold regularized multi-task feature selection (M2TFS)~\cite{Jie2013}, used by Jie \etal, also trains a sparse linear regressor but is specifically designed for selecting nodal features~\cite{Jie2014a}. M2TFS adds a second regularization term, based on the graph Laplacian, that leverages the network structure and encourages features from neighbouring nodes to be selected together. Additionally, M2TFS generalizes the feature selection model to extract multiple types of features at each node, encouraging features from the same node to be selected together.

Finally, Ball \etal\ used Boruta feature selection~\cite{Kursa2010}, which, instead of model weights, uses the frequency of feature selection of a random forest classifier~\cite{Ball2016}. First, a null distribution of the frequency selection of each feature is created by training a random forest on data with permuted class labels. Then, a second random forest is trained on the data with the correct class labels and only those features selected by the random forest significantly more often than chance are kept for the final classifier. 

As is the case for the filter type feature selectors, different wrapper based feature selectors have their own advantages and disadvantages and so it is not obvious which wrapper type approach is best for features from connectome data. Note, however, that of the wrapper type feature selectors used by studies listed in Table~\ref{tab:depthReportTable}, only M2TFS and DSM leveraged the network structure of the connectome data. Ideally, a feature selector should utilize the connectome topology to filter features intelligently. Also, an ideal wrapper type feature selector should employ a model that matches the type of model used in the actual learning algorithm, since different learning models may be sensitive to different types of features and different patterns of feature values. Even with matched model types, all wrapper type feature selectors have the drawback that the feature selection model is divorced from the final learning model. This adds additional training time and also means that the set of selected features may not be optimal for the final model. In Section~\ref{sec:learningModels} we discuss embedded feature selectors which take the form of regularization terms in a learning model's objective function.

\subsubsection{Dimensionality Reduction Techniques}
\label{sec:dimensionalityReduction}
Rather than filtering out certain features entirely, dimensionality reduction techniques extract salient information by transforming the input features to a lower dimensional space in which the cardinal directions covary with important factors in the data. For instance, five works from Table~\ref{tab:depthReportTable} employed principal component analysis (PCA) which uses eigenvalue decomposition to transform the features onto an orthogonal basis representing the major modes of variation in the feature space~\cite{Ziv2013,Brown2015,Mitra2016,Robinson2010,Prasad2014}. Note that Ziv \etal~\cite{Ziv2013} and Mitra \etal~\cite{Mitra2016} both applied PCA after applying an initial feature selection step to reduce the dimensionality of the features even further. One drawback of PCA is that it assumes a Gaussian model over the data, so data that is distributed over more complex manifolds will not be represented correctly.

In contrast, local linear embedding (LLE)~\cite{Roweis2000}, used by Shen \etal~\cite{Shen2010}, makes no assumption about the shape or topology of the feature space manifold other than that it can be approximated as a linear function at each point. LLE provides a mapping between the high dimensional manifold and a flattened, lower dimensional space that captures the dominant modes of variation. Shen \etal\ used the maximum likelihood estimator~\cite{Levina2004} to determine the intrinsic dimensionality of the feature space manifold for which to map to. After regularizing each correlation matrix to be semi-positive definite (SPD) using GLASSO, Qui \etal\ reduced dimensionality using their proposed SPD-LLE which extends LLE by restricting the manifold to the SPD Riemannian space.

Given the matrix of feature row vectors for each connectome, $X \in \mathbb{R}^{N \times M}$, Munsell \etal\ transformed their features using a linear kernel based on the Gram matrix of $X$~\cite{Munsell2015}. The output of the linear kernel is a new feature matrix $\hat{X} = X^T X$, in which the $ith$ feature of the $jth$ connectome is the similarity between the original feature vectors of the $ith$ and $jth$ connectomes, computed as $\hat{X}_{ij} = X_i \cdot X_j$. Thus, each connectome is described in terms of its similarity to other connectomes. The linear kernel brings the number of features, $M$, down (or up) to $N$, which can be especially adventageous if the number of features is far greater than the number of instances (\ie\ $M \gg N$) which, as mentioned above, is common in neuroimaging applications. A variety of other studies also used kernels, specifically in the context of kernel support vector machines (SVM) and so these approaches are discussed below in Section~\ref{sec:learningModels}.

In non-negative matrix factorization (NMF)~\cite{Lee1999}, used for dimensionality reduction by Anderson \etal~\cite{Anderson2014}, transformed features, $\hat{X} \in \mathbb{R}^{N \times \hat{M}}$, represent weights of a basis set (dictionary) of learned, high-level, non-negative features $D \in \mathbb{R}^{\hat{M} \times M}$ (where $\hat{M}$ is the number of output features) . The original feature matrix is decomposed as $X = \hat{X}D + \varepsilon$ and $\varepsilon$ is a residual matrix that is minimized. Due to the linearity of the model, the non-negativity constraint acts as a sparsity term, encouraging small entries in $D$ and $\hat{X}$ to go to zero. Thus, in the case that $X$ represents edge or node features for each connectome, $D$ represents a basis of consistent subnetworks. Note that each subnetwork may be internally connected or disconnected, depending on which edges or nodes covary consistently.

A general drawback of using dimensionality reduction techniques is that they are not bijective functions, and so information useful for visualizing discriminative features (in particular edges, nodes and subnetworks) can be lost. This issue is especially important for connectome data since basic edge and node features have an intrinsic anatomical location (and often a known or hypothesized anatomical function) which makes visualization of learned models very informative.

\subsection{Learning Models}
\label{sec:learningModels}

In this section, we summarize the wide variety of models that have been used for machine learning on connectome data, including supervised prediction models (Section~\ref{sec:predictionModels}) such as linear prediction models (Section~\ref{sec:linearModels}), kernel based models (Section~\ref{sec:graphKernels}), probabilistic models (Section~\ref{sec:probabilisticModels}), ensembles (Section~\ref{sec:ensembles}) and unsupervised models (Section~\ref{sec:unsupervisedModels}).

\subsubsection{Prediction Models}
\label{sec:predictionModels}
As mentioned above, the majority of the studies discussed in this review performed supervised outcome prediction. Let $X \in \mathbb{R}^{N \times M}$, be a matrix containing feature row vectors for each connectome and let $\mathbf{y} \in \mathbb{R}^{N \times 1}$ be the associated ground truth labels (assuming a single label per connectome). A prediction model can be represented as a function, $f(X_i;\Theta)$, where $X_i \in  \mathbb{R}^{1 \times M}$ is a row vector of $X$, containing features of the $ith$ connectome and $\Theta$ is a set of learnable model parameters.

The learning step of many predictive machine learning models can then be expressed mathematically as an optimization of an objective function,

\begin{equation}
\Theta^* = {\underset {\Theta}{\text{argmin}}}\: \mathcal{L} (\mathbf{y}, \hat{\mathbf{y}} = \mathbf{f}(X; \Theta) ) + R(\Theta),
\label{eq:learning}
\end{equation}
\noindent subject to some (possibly empty) set of constraints, $\mathcal{C}$, where $\mathcal{L}$ is the loss function (or data term),
$\hat{\mathbf{y}} \in \mathbb{R}^{N \times 1}$ is the vector of output labels predicted by $\mathbf{f} = [f(X_1;\Theta), \ldots, f(X_N;\Theta)]^T$, and $R$ is the regularization function (sometimes called the prior as it often encodes prior information). By optimizing over Eq.~\ref{eq:learning}, we find a regularized and potentially constrained set of model parameters, $\Theta^*$, that minimize the loss between the labels predicted by the model, $\hat{\mathbf{y}}$, and the ground truth labels, $\mathbf{y}$.

\subsubsection{Linear Prediction Models}
\label{sec:linearModels}
The most basic prediction model, linear regression, has been used for connectome data with different regularization terms~\cite{Brown2016,Park2015}. For linear regression, the set of learnable weights comprises a single vector of weights $\Theta = \{ \mathbf{w} \}$, where, $\mathbf{w} \in \mathbb{R}^{M \times 1}$, $f(X_i;\mathbf{w}) = X_i \cdot\mathbf{w}$ and $\mathcal{L}(\mathbf{y}, \mathbf{f}(X;\mathbf{w})) = ||\mathbf{y} - X\mathbf{w}||^2_2$. Note that $f$ may contain a bias term (\ie\ y-intercept) but for simplicity, here and throughout the rest of the paper, we assume that this bias term is included as an extra weight in $\mathbf{w}$. Thus, in the case that a bias term is included, we have $\mathbf{w} \in \mathbb{R}^{(M+1) \times 1}$ and $X \in \mathbb{R}^{N \times (M+1)}$ contains an additional, final column of all 1's.

The LASSO model, used by Vanderweyen \etal~\cite{vanderweyen2013} and Wee \etal~\cite{Wee2016} for feature selection is simply linear regression with the $\ell_1$ regularization term, $R(\mathbf{w})=\lambda_{\ell_1}||\mathbf{w}||_1$, where $\lambda_{\ell_1}$ is a user specified weight (\ie\ hyper-parameter) that determines the strength of regularization. Note that in this case, $R(\mathbf{w})$ can be considered an embedded feature selector, as it helps to determine the set of features used in the final prediction model. 

As mentioned above, the objective function for ElasticNet is the same as that for LASSO but with regularization function, $R(\mathbf{w})=\lambda_{\ell_1}||\mathbf{w}||_1 + \lambda_{\ell_2}||\mathbf{w}||^2_2$, where $\lambda_{\ell_2}$ is the weight on the smoothness term. Typically, a unity-sum constraint, $\lambda_{\ell_1} = 1 - \lambda_{\ell_2}$, is also enforced. (Note that if only the $\ell_2$ term is present, it is called a Tikhonov regularization, and the entire model is called a ridge regression model~\cite{golub1999tikhonov,hoerl1970ridge}).

Similarly, Brown \etal\ extended LASSO for their prediction model with two additional terms that encouraged features to be selected 1) from edges on the connectome backbone (\ie\ edges with high signal to noise ratio) and 2) from edges connected to one another (\ie\ sharing the same nodes)~\cite{Brown2016}. They found that this model out-performed both LASSO and ElasticNet on their perterm infant dataset.

Park \etal\ instead employed partial least-square regression (PLSR)~\cite{Krishnan2011} which combines linear regression with PCA~\cite{Park2015}. Rather than minimizing prediction error between ground truth and predicted labels, PLSR aims to find a set of $L=\text{rank}(X)$ orthonormal latent variables, $\{ \mathbf{t}_{\ell} \in \mathbb{R}^{N \times 1} : \ell \in [1, L] \}$, that covary with the labels. In particular, for a set of $L$ weight vectors, $\Theta = \{ \mathbf{w}_{\ell} : \ell \in \{1,2, \ldots, L\} \}$, and sample covariance function, $\text{cov}(\cdot)$, the loss is defined as $\mathcal{L}(\mathbf{y}, \mathbf{f}(X;\Theta)) = - \sum^L_{\ell = 1}{\text{cov}(\mathbf{t}_{\ell}, \mathbf{y}) }$, subject to the constraints,
\begin{align}
C_0: \mathbf{t}_{\ell} = X\mathbf{w}_{\ell} \\
 C_1: if \: \ell \neq m \: \: \text{then} \: \mathbf{t}_{\ell} \mathbf{t}_{m}^T = 0 \\
\text{else} \: \mathbf{t}_{\ell} \mathbf{t}_{\ell}^T = 1
\end{align}
which ensure orthonormality of latent variables. Krishnan \etal\ presented an iterative algorithm to optimize this objective function that involves use of the singular value decomposition~\cite{Krishnan2011}. The advantage of PLSR is that dimensionality reduction and predictive model learning are performed in a single stage.

Linear discriminant analysis (LDA)~\cite{Friedman1989} (sometimes referred to as Fisher linear discriminant analysis) is another classical linear model that has been used with connectome data for supervised prediction of categorical labels~\cite{Alnæs2015,Kaufmann2016,Robinson2010,chen2011,Caeyenberghs2013}. 
 In LDA, the feature space distribution of samples (\ie\ of connectomes) in each class is assumed to be Gaussian, and the goal is to find a set of weights, $w$, that minimizes intra-class variance and maximizes inter-class variance. In two-class LDA, given sample class means, $\mathbf{\mu}_0$ and $\mathbf{\mu}_1$, and a single sample covariance matrix, assumed (via the homoscedasticity assumption) to be shared by both classes, $\Sigma = \frac{1}{2}(\Sigma_0 + \Sigma_1)$, the LDA loss function is defined as, 
\begin{equation}
\mathcal{L}(w) = \frac{ 2 \mathbf{w}^T (\Sigma) \mathbf{w}  }{(\mathbf{w} \cdot (\mu_1 - \mu_0))^2}.
\end{equation}

\noindent The weights can then be computed analytically as, $\mathbf{w} = \Sigma^{-1}(\mathbf{\mu}_1 - \mathbf{\mu}_0)$. The binary prediction model takes the form of a linear boundary (\ie\ hyperplane) with the prediction function, $f:\mathbb{R}^{M} \rightarrow \{0,1\}$, defined as,

\begin{equation}
f(X_i; \mathbf{w}) = 
\begin{cases}
1, & \text{if } X_i \cdot \mathbf{w} > \frac{1}{2}(\mathbf{\mu_1}^T \Sigma^{-1} \mathbf{\mu_1} - \mathbf{\mu_0}^T \Sigma^{-1} \mathbf{\mu_0}) \\
0, & \text{otherwise.}
\end{cases}
\end{equation}

\noindent Maximum uncertainty linear discriminant analysis (MLDA)~\cite{Thomaz2006}, used by Robinson \etal\ and regularized LDA (rLDA)~\cite{Friedman1989}, used by Aln\ae s \etal\ and Kaufmann \etal, are both variants of LDA  in which the covariance matrix is estimated more conservatively using covariance shrinkage~\cite{Alnæs2015,Kaufmann2016,Robinson2010}.

\subsubsection{SVM and Kernels}
\label{sec:graphKernels}
Of papers examined by this review, SVM was the most widely used prediction model (\nPapersUsedSVM~papers used SVM)~\cite{Brown2015,Cuingnet2011,Zhu2014,Ziv2013,Solmaz2016,Li2012,Guo2012,Ng2016,Prasad2014,Prasad2015,Wee2011,
Zhu2014a,Moyer2015,Wee2013,Khazaee2015a,Khazaee2015,Wee2013a,vanderweyen2013,Pruett2015,Rosa2015,Sacchet2015,
Galvis2016}. Rather than explicitly modelling the feature distributions of each class like LDA, SVM defines a class boundary hyperplane that is informed only by the samples nearest to it, known as support vectors. Formally, given binary ground truth labels, $y_i \in \{-1, 1\}$, SVM aims to find a hyperplane, defined by weights, $\mathbf{w} \in \mathbb{R}^{M \times 1}$, which satisfy the constraints: 
\begin{equation}
C_i: y_i (X_i \cdot \mathbf{w}) > 1 - \xi_i, \forall i \in [1,N],
\end{equation}
\noindent where $\{\xi_i : i \in [1,N]\}$ is a set of slack variables which represent violations of the constraints. The goal of the data term is only to minimize the slack variables, $\mathcal{L}(\mathbf{y}; \xi)  = \lambda_{\xi} \sum^N_{i=1}{\xi_i}$. Because this problem is ill posed, typically an $\ell_2$ regularization term is imposed (\ie\ $R(\mathbf{w})=||\mathbf{w}||^2_2$ ), though some studies (\eg\ Rosa \etal~\cite{Rosa2013}) used an $\ell_1$ term instead to promote sparse weights.

Cuingnet \etal\ and Li \etal\ both incorporated a Laplacian based regularizer into SVM to promote spatial smoothness in edge features selection weights accross the network~\cite{Cuingnet2011,Li2012}. Given the graph Laplacian, $\mathcal{L}$, and regularization strength, $\lambda_{\beta}$, this regularization term takes the form, 
\begin{equation}
R(\mathbf{w}) = ||e^{\frac{1}{2} \lambda_{\beta} L} \mathbf{w}||^2, 
\end{equation}
\noindent which resembles a heat flow kernel, diffusing the weights across the network. This diffusion encourages topologically local edges of the network to share strong edge weights and reduces the weight on lone, weakly connected edges.

Prasad \etal\ used SVM as part of a simulated annealing framework to find the best set of ROIs (\ie\ node definitions)~\cite{Prasad2014}. In their case, each ROI represented a non-empty subset of the cortical regions parcellated, in each scan, by an atlas based segmentation. The simulated annealing procedure optimized the ROIs by randomly permuting the parcellated regions in each ROI, quickly at first and then with increasingly smaller pertubations in later iterations. The SVM classification accuracy in each iteration was used to rate the quality of that candidate set of ROIs. Such simulated annealing algorithms are typically computationally intensive, but can provide good quality approximate solutions to NP Hard problems~\cite{eglese1990simulated}.

Many approaches have used kernel SVM which enables the normally linear SVM to model a (potentially highly) non-linear decision boundary ~\cite{Arbabshirani2013,Dodero2015,Jie2014,Guo2012,Jin2015,Wee2016,Vega-Pons2014,Takerkart2012,Ball2016,Wang2014,Mokhtari2012,Jie2014b,Jie2014a,Arbabshirani2013,Shahnazian2012,
Kamiya2015,Bassett2012}. Kernel SVM modifies the standard SVM constraint, instead requiring:
\begin{equation}
C_i: y_i (\Phi(X_i) \cdot \mathbf{w}) > 1 - \xi_i, \forall i \in [1,N],
\end{equation}
\noindent where $\Phi: \mathbb{R} \rightarrow \upchi $ is a function that transforms feature vectors to some new space, $\upchi$. The kernel function is then defined as $K_{ij}(X_i, X_j) = \Phi(X_i)^T \Phi(X_i)$ and represents an inner product between $X_i$ and $X_j$ in the new space, $\upchi$. Note that if $\Phi(X_i)=X_i$ then $K$ is the linear kernel, and kernel SVM becomes equivalent to standard SVM.

A variety of kernels have been explored for use on connectome data. One of the most common kernels, used by four different studies in Table~\ref{tab:depthReportTable}, is the radial basis function (RBF) kernel~\cite{Guo2012,Ball2016,Arbabshirani2013,Kamiya2015}. The RBF kernel is defined as, 
\begin{equation}
K_{ij}(X_i, X_j; \sigma) = e^{-\frac{1}{2 \sigma^2} ||X_i - X_j||^2},
\end{equation} 
\noindent where $\sigma$ is a hyperparameter defining the isotropic standard deviation of the kernel. Note however that the RBF treats each feature equally which may not be ideal when comparing two connectomes, especially if prior information is known about how the topology of the connectomes may vary between classes.

Other approaches have used kernels that explicitly compare different topological features between two graphs including features of random walks, subnetworks and subtrees~\cite{Mokhtari2012}. For instance, Mokhtari \etal, Jie \etal, Vega-Pons \etal\, Fei \etal\ and Wang \etal\ all used the Weisfeiler-Lehman (WL) subtree kernel~\cite{Shervashidze2010}, which efficiently counts the number of matching subtree patterns in a pair of networks~\cite{Mokhtari2012,Jie2014,Vega-Pons2014,Fei2014,Wang2014}. Note that the WL kernel can be applied to any two input networks and makes no assumption about correspondence between nodes in the two networks.

A random walk on a graph refers to a sequence of nodes in which the next node is sampled randomly based on the distribution of weights of the edges connected to the current node. Shahnazian \etal\ used a kernel that measures the similarity between simultaneous random walks, each performed on one of two connectomes~\cite{Vishwanathan2007,Shahnazian2012}. In a comparative study, Shahnazian \etal\ found that the random walk kernel outperformed the WL subtree kernel~\cite{Shervashidze2010} and a kernel based on comparing subnetworks~\cite{Shervashidze2008}, in classifying resting from attention states.

Takerkart \etal\ proposed three kernels that compute similarity in connectome function, ROI locations and ROI adjacency and combined them into a single kernel~\cite{Takerkart2012}. In their approach, each connectome was constructed such that edges were defined by spatial adjacency between neighbouring ROIs, nodal positions were defined by ROI centroids and averaged fMRI time series at each ROI defined nodal signals. Given two connectomes, $G(E,V)$ and $G'(E',V')$, the combined kernel is defined as, 

\begin{equation}
K(G,G') = \sum_{e_{p,q} \in E} \sum_{e_{p',q'} \in E'} k_s(e_{p,q},e_{p',q'}) \cdot k_g(e_{p,q},e_{p',q'}) \cdot k_f(e_{p,q}, e_{p',q'}),
\label{eq:spatialTopoKernel}
\end{equation}

\noindent where $k_s$ is 1 if both $e_{p,q}$ and $e_{p',q'}$ exist (\ie\ nodes $p$ and $q$ represent spatially adjacent ROIs in $G$ and nodes $p'$ and $q'$ represent spatially adjacent ROIs in $G'$ ) and 0 otherwise, $k_g$ computes the similarity between positions of nodes $p$ and $p'$ and nodes $q$ and $q'$, and $k_f$ computes similarity between fMRI signals, also between the two pairs of nodes. (Note that there is no special correspondence between $p$ and $p'$ or between $q$ and $q'$ and that Eq.~\ref{eq:spatialTopoKernel} sums over every possible pair of edges in the two connectomes.)

Other than the random walk kernel, none of the graph kernels discussed above make an assumption of correspondence between nodes in different connectomes and so are all applicable to cases where connectomes are not constructed using corresponding landmarks or atlas regions. Coversley, for connectomes that are constructed with consistent ROI across the dataset, these kernel can not leverage the intrinsic correspondence that exists between nodes and edges, and so may be at a disadvantage compared to other measures of connectome similarity.

\subsubsection{Probabilistic Models}
\label{sec:probabilisticModels}
Logistic regression~\cite{hosmer2004applied} is a probabilistic linear model that has been used for prediction of outcomes using connectome data but is only applicable when the class labels are binary. Logistic regression uses the same prediction function as the standard linear model, $f(X_i;\mathbf{w}) = X_i \cdot \mathbf{w}$, but uses a different loss function that estimates the negative log likelihood of the labels, given the features,
\begin{equation}
\mathcal{L}(\mathbf{y}, \mathbf{f}(X;\mathbf{w})) = \frac{1}{N} \sum^N_{i=1}{log(1+ e^{-y_i (X_i \cdot \mathbf{w})})}.
\end{equation}

\noindent Of the papers in this review, three used logistic regression and two of those used sparse logistic regression (SLR)~\cite{Ryali2010}, which is logisitic regression with the inclusion of an $\ell_1$ regularization term~\cite{Ng2012,Guo2012,Zhan2015}. Additionally, Ng \etal\ also included a regularization term based on the graph Laplacian~\cite{Ng2012}. Their regularization function, $R(\mathbf{w})=\lambda ||\mathbf{w}||_1 + (1 - \lambda)\mathbf{w}^T L \mathbf{w}$, which combines $\ell_1$ based sparsity and Laplacian based spatial regularization, is called a GraphNet regularizer~\cite{Grosenick2013}.

Varoquax \etal\ classified stroke patients from normal controls using a statistical model of the covariance matrices of the controls only~\cite{Varoquaux2010}. To do this, they defined a generalized Gaussian distribution over the Riemannian space of covariance matrices. To address the HDSSS problem, they used a fixed variance, greatly reducing the number of model parameters. Instead, Chung \etal\ used a Gaussian na{\"i}ve Bayes classifier~\cite{zhang2004optimality} which models the probability of each class given the input features, which are assumed to be independent and normally distributed~\cite{Chung2016}.

Moyer \etal\ also employed a na{\"i}ve Bayes classifier but modelled the feature distribution of each class as distribution of connectomes with assumed community structure using a mixed membership stochastic blockmodel (MMSB)~\cite{Moyer2015,karrer2011stochastic}. The MMSB assumes that the membership of each node into $K$ possibly overlapping communities (\ie\ subnetworks) is modelled by a multinomial distribution and that the connectivity (\ie\ tract count) between any pair of nodes, given the their community memberships, is modelled by a Poisson distribution. An MMSB was learned for each class and then each unseen connectome was assigned the class associated with the MMSB to which that connectome fit with highest probability.

Sweet \etal\ predicted the locations of epilepsy in the brain by constructing a model of the probability of each ROI in a functional connectome being healthy or epileptic~\cite{Sweet2013}. To predict epileptic regions from observed functional connections, they first introduced latent variables representing the true (but unknown) functional connectivity in normal and abnormal connectomes. Observed functional connectivities of each edge were assumed to be noisy measurements of the true connectivities. They then modelled the probability of an edge connectivity in an abnormal scan, given the connectivity of the same edge in a normal scan and given the health of the two end-point ROIs (\ie\ either healthy or epileptic regions). Finally, the probability of given ROI being epileptic in an unseen test connectome was computed as a marginal posterior probability, based on the connectome functional edge connectivities and the learned model parameters.

Generally, probablistic models are riggorously motivated and handle uncertainty in features and labels explicitly which can provide additional information about model predictions. However, this extra information tends to come at the price of (potentially many) additional parameters that must be learned (\eg\ $M \times M$ covariance matrices) and/or increased number of samples required per model parameter (\eg\ logistic regression requires at least 10 samples per independent parameter versus only 5 for ordinary least squares used for linear regression~\cite{peduzzi1996simulation}).

\subsubsection{Ensembles and Stacked Models}
\label{sec:ensembles}
A variety of papers on the application of machine learning to connectome data used ensemble learning models such as random forests~\cite{Breiman2001}, boosting models~\cite{Freund1995}, and deep learning models~\cite{LeCun2015}, which combine multiple so-called weak learner models. Six works used random forests classifiers~\cite{Anderson2014,Arbabshirani2013,Mitra2016,Guo2012,Richiardi2011,Richiardi2012}. Random forests are ensembles of decision tree classifiers (often learned in parallel) in which each decision tree (\ie\ weak learner) is given a random subset of features and the final prediction is often decided by majority voting among trees~\cite{Breiman2001}. Each tree is trained as a recursive splitting of the dataset at each tree node by the single feature and associated feature threshold value that best splits the data at that node. Functional trees~\cite{Gama2004}, used by Richiardi \etal\, generalize decision trees by allowing multiple features to define a decision boundary at each node~\cite{Richiardi2011,Richiardi2012}. Also, instead of randomly splitting features, Richiardi \etal\ split the fMRI signals into four different frequency subbands and trained each tree on connectome features derived from a unique subband.

Rather than training weak learner classifier in parallel (like random forests), boosting algorithms train each classifier sequentially, using the classification accuracies of previously trained classifiers to inform the model parameters of the next classifier. Adaboost~\cite{Freund1995}, a boosting algorithm used by Pariyadath \etal~\cite{Pariyadath2014}, forms a final ensemble classifier by taking a linear combination of the trained classifiers, each weighted by its classification accuracy. Pariyadath \etal\ used SVM models for each weak learner classifier in the Adaboost ensemble. They showed that this ensemble could predict whether participants of the study were smokers or not by examining functional subnetwork-based features of participants' connectomes~\cite{Pariyadath2014}.

While the k nearest neighbours (kNN) classifier, used by Arbabshirani \etal, is not an ensemble method, it does use majority voting to decide the class of each test sample (like random forests)~\cite{Arbabshirani2013}. However, whereas voting is performed by multiple decision tree classifiers in each random forest, in kNN, voting is performed by the k training samples with the most similar feature vectors where each votes with its own class label.

A stacked model is another kind of ensemble in which the outputs of each weak learner are provided as inputs of the next weak learner. Deep learning models are stacked models that posess many layers (\eg\ more than 2) and have recently become very popular in certain domains due to their improved performance over other models on a variety of tasks (\eg\ AlexNet for natural image classification~\cite{Krizhevsky2012})~\cite{LeCun2015}. While deep models have been applied readily to standard grid-like images of the brain~\cite{Yoo2014,Yang2014,Liu2014deep,Li2014,Brosch2013,Suk2015,dvorak2015structured} and even dMRI and fMRI data (\eg\ Plis \etal~\cite{Plis2014}), their use for connectome data has been much more limited. This is likely because deep models often posses thousands or millions of learnable parameters which require large datasets to train on~\cite{LeCun2015}. As was mentioned in Section~\ref{sec:outcomePrediction}, large open MRI connectome datasets have only recently become available and they currently contain hundreds or thousands of scans, but not millions.

Nevertheless, Munsell \etal\ used a deep stacked auto-encoder~\cite{Hinton2006} to learn a set of relevant edge features to select from each connectome~\cite{Munsell2015}. Breifly, each auto-encoder layer, comprising a complete bipartite graph of inputs and outputs (also known as a fully-connected layer), aims to encode its input into a lower dimensional space with minimal loss of information (\ie\  between original input and decoded encoded input). The auto-encoder layers are then stacked such that each layer's output code  becomes the input of a smaller (\ie\ fewer weights) auto-encoder, where the values of each output are first passed through a non-linear activation function. The result is a model capable of learning highly non-linear relationships in the data. While Munsell \etal\ found that a linear feature selector outperformed the deep model, their dataset only contain 118 images, and so the deep model likely over-fit to the training set and was not able to generalize well.

One problem with fully-connected layers is that, for $M$ inputs and $M'$ outputs, they require learning $\mathcal{O}(MM')$ parameters, which is often a large number. Kawahara and Brown \etal\ proposed a deep CNN called BrainNetCNN with convolutional layers designed specifically for connectome data that leveraged the topology of the input brain networks~\cite{kawahara2016}. Each of their proposed convolutional layers required learning only $\mathcal{O}(M)$ parameters. For models with the same number of parameters, they found that their BrainNetCNN model outperformed a model with only fully-connected layers. In order to address the challenges of performing deep learning on a relatively small dataset (168 scans), Kawahara and Brown \etal\ also augmented their dataset by generating synthetic connectomes via SMOTE~\cite{chawla2002smote}.

Generally, ensembles and stacked learning methods enable highly flexible, non-linear models to be trained but require large datasets and greater computational resources to do so. Given that the amount of available connectome data seems to be increasing (Fig.~\ref{fig:numScansVsYear}) and that computational power continues to follow Moore's law, we expect ensembles and deep neural networks in particular to become much more widely studied for applications to brain networks.

\subsubsection{Unsupervised Models and Dictionary Learning Models}
\label{sec:unsupervisedModels}

A small subset of the papers in Table~\ref{tab:depthReportTable} used unsupervised learning models on their connectome data, including NMF and clustering algorithms like k-means. The k-means algorithm groups the input data into k clusters by iteratively finding the feature space mean of each cluster and then reassigning each sample to its nearest cluster based on the new mean~\cite{forgy1965cluster}. Shen \etal\ used k-means\footnote{Shen \etal\ call their clustering algorithm c-means, which is a fuzzy version of k-means, but the algorithm they describe is actually k-means~\cite{Shen2010}.} clustering to split the data into two groups after reducing the dimensionality of each connectome via LLE~\cite{Shen2010}. They found that, in the reduced dimensionality space, the connectome clusters match the ground-truth schizophrenia/control class labels well.

Ashikh \etal\ instead used the ordering points to identify clustering structure (OPTICS) algorithm~\cite{Ankerst1999}, which requires, as input, the minimum number of points (in this case connectomes) per cluster in lieu of the final number of custers (as k-means does)~\cite{Ashikh2015}. Rather than explicitly determining clusters, OPTICS uses the distance between points to induce a density based ordering on all points in the dataset from which clusters can be inferred. Despite using a Euclidean distance metric between edge features of each connectome, with no dimensionality reduction (aside from filtering via a t-test), Ashikh \etal\ reported that the OPTICS algorithm was able to group connectomes into three clusters which perfectly matched the early MCI, late MCI and AD class label groups in their dataset.

Li \etal\ also used k-means clustering but only as a way of algorithmically assigning binary labels to each unlabelled connectome before running Fisher discrimination dictionary learning (FDDL)~\cite{Yang2011,Li2014b}. In FDDL, the goal is to learn a dictionary, $D \in \mathbb{R}^{M \times K}$, of $K$ atoms (each with $M$ features) and a sparse encoding of the dataset $W \in \mathbb{R}^{K \times N}$, such that,
\begin{align}
\label{eq:dictionaryLearning}
\{D^* , W^* \} = {\underset {D,W}{\text{argmin}}}\: \mathcal{L} (X; D,W) + R(W), \\
\text{and}\:\: ||D_m||_2 = 1,\: \forall i \in [1,M].
\end{align}

\noindent The loss function, $\mathcal{L} (X; D,W)$, encourages the code matrix, $W$, to encode the original data when projected onto the dictionary, $D$, and also encourages different subsets of the rows of $D$ (\ie\ sub-dictionaries) to encode class specific information. The regularization term, $R(W)$, encourages the codes in $W$ to be sparse and discriminative (via the Fisher discriminant criterion~\cite{Yang2011}). For full details about the loss and regularization functions, see Yang \etal~\cite{Yang2011}. Applying this method, Li \etal\ found two (of the $K=16$, in their case) learned connectome subnetworks, defined by columns in $D$, that could differentiate between PTSD and control in $80\%$ of their dataset.

NMF~\cite{Lee1999}, used by Ghanbari \etal\ (and by Anderson \etal\ for dimensionality reduction, see Section~\ref{sec:dimensionalityReduction}), is similar to dictionary learning in that the goal is to learn a dictionary, $D \in \mathbb{R}^{M \times K}$, and a code matrix, $W \in \mathbb{R}^{K \times N}$, of $X$ such that the loss $\mathcal{L}(X; D,W) = ||X^T - DW||^2_F$ is minimized (where $||\cdot||_F$ is the Frobenius norm)~\cite{Ghanbari2014}. This is the same as saying that $W$ projected onto $D$ should fit closely to the data $X$, which was one of the goals of FDDL~\cite{Yang2011}. However, in NMF, a constraint is added that $D_{ij} \geq 0, \forall i,j$ and also it is assumed that $W=D^T X^T$ (\ie\ $W$ is the projection of $X$ onto $D$) and so the loss becomes $\mathcal{L}(X; D) = ||X^T - DD^TX^T||^2_F$. Like in FDDL, Ghanbari \etal\ also split the dictionary into sub-dictionaries, $D=[\hat{D}, \breve{D}, \tilde{D}]$, with different objectives/penalties for each. However, instead of each sub-dictionary being associated with a class, they encouraged the sub-dictionaries to be class discriminative, age regressive and reconstructive with three separate regularization terms in $R(\hat{D}, \breve{D}, \tilde{D})$.

An advantage of both dictionary learning methods is that, assuming $X$ represents edge features for each connectome (which is true for both Li \etal\ and Ghanbari \etal) then the atoms of $D$ (or topics as they are referred to in Ghanbari \etal) represent sparse connectome subnetworks which have intrinsic meaning, defined by their parent subdictionary's loss/regularization function, and which can be easily visualized to show which parts of the brain are involved. However, a major disadvantage of both FDDL and NMF is that the number of atoms must be selected \emph{a-priori}, and there may be no clear, principled way to choose this.

Whereas dictionary learning approaches find subnetworks important for groups of connectomes, the stable overlapping replicator dynamics (SORD), proposed by Yoldemir \etal\ is designed to learn important subnetworks within a single connectome (though they show how the technique can be applied to a group of functional connectomes by concatenating the fMRI signals across time)~\cite{Yoldemir2015}. A strongly connected subnetwork of nodes can be identified by node weights, $\mathbf{w}$, by optimizing:
\begin{align}
\label{eq:SORD}
\mathbf{w}^* = {\underset {\mathbf{w}}{\text{argmax}}}\: \mathbf{w}^T A \mathbf{w}\\
\text{such that}\: \: ||\mathbf{w}||_1 = 1 \:\: \text{and} \:\: \mathbf{w} \geq 1,
\end{align}

\noindent where $A \in \mathbb{R}^{|V| \times |V|}$ is the connectome adjacency matrix. SORD allows multiple, possibly overlapping subnetworks to be identified by iteratively adding a new node and new edges to the connectome (\ie\ new row and column in $A$) after each identification of a subnetwork. The edges of the new node are engineered to destabilizes the previous solution, forcing a new subnetwork to be found. They used a bootstrapping technique to ensure the stability of edges found in each subnetwork. Yoldemir \etal\ also extended SORD to coupled SORD (CSORD) that can identify subnetworks that are both structurally and functionally well connected~\cite{Yoldemir2015coupled}. Given adjacency matrices of structural, $A^s$, and functional, $A^f$, connectomes of the same brain, and two vectors of nodal weights, $\mathbf{w}^a$, and $\mathbf{w}^b$, the subnetwork can be identified by optimizing:
\begin{align}
\label{eq:CSORD}
\{ \mathbf{w}^{a*}, \mathbf{w}^{b*}\} = {\underset {\mathbf{w}^a, \mathbf{w}^b}{\text{argmax}}}\: (\mathbf{w}^a)^T A^s \mathbf{w}^b + (\mathbf{w}^b)^T A^f \mathbf{w}^a\\
\text{such that}\: \: ||\mathbf{w}^a||_1 = 1, \:\: \mathbf{w}^a \geq 1, \:\: ||\mathbf{w}^b||_1 = 1 \:\: \text{and} \:\: \mathbf{w}^b \geq 1,
\end{align}
\noindent for which $\mathbf{w}^a$, and $\mathbf{w}^b$ tend to converge for dense networks, giving a single multi-modal subnetwork (per iteration).

Finally, Arslan \etal\ also proposed a method to group nodes, but instead of replicator dynamics, they employed clustering on an eigenvector representation of the input connectomes. In particular, given a cortical surface mesh with edge weights defined by functional correlation (between spatially local pairs of mesh nodes only), the task was to group mesh nodes into larger ROIs with consistent fMRI signals. To do so, they computed the eigenvalue (spectral) decomposition of the mesh grid's Laplacian matrix and extracted a reduced dimensionality representation of each mesh vertex (by retaining the eigenvectors corresponding to the top eigenvectors). For a single subject parcellation, they applied k-means to the reduced dimensionality mesh vertices, resulting in k ROIs. For a group-wise parcellation, they first created a single mesh graph with correspondence edges between spatially local nodes in different graphs, weighted by the correlations between nodal connectivity patterns. Then, they performed an eigenvalue decomposition on the combined group graph, and again applied k-means to find clusters of nodes across different mesh graphs. The final set of ROIs was computed by majority voting across subjects.

Unsupervised learning methods have provided ways to explore the structure within and between connectomes. Given the complexity of the brain and amount of variation within a single brain across time, or between different subjects' brains across the population, we expect that the use of unsupervised learning for further exploration of the human connectome will continue to expand.

\subsection{Validation}
\label{sec:validation}
A wide variety of measures have been used to validate the different machine learning models. By far the most popular measure was standard prediction accuracy (ACC), which was used in \nPapersUsedAcc\ of \nPapers\ papers. Note however that for imbalanced data (Section~\ref{sec:classImbalance}), accuracy may over-estimate the performance of the model. Thus, nearly half of the papers in this study also reported sensitivty (SN, also known as recall) and specificity (SP), which convey how the model performs on positive and negative instances, respectively (assuming 2-class prediction). Other reported validation measures included positive predictive value (PPV, also known as precision), negative predictive value (NPV), F-score (FS) and the area under the receiver-operator characteristic curve (AUC). Some papers have reported variations on these measures including corrected PPV (cPPV) and corrected NPV (cNPV), which take disease (\ie\ positive class) prevalence in the overall population into account~\cite{altman1994statistics}, and balanced accuracy (BAC), which averages accuracy computed over each class separately in order to remove the effect of class imbalance~\cite{brodersen2010balanced}. Wee \etal\ additionally reported Youden's index (YDI), defined as $SN + SP -1$, which, similar to balanced accuracy, conveys the sensitivity and specificitiy of the predictor in balanced way~\cite{Wee2012}.

For the minority of studies that performed regression instead of classification, a different set of validation measures were used, including mean absolute error (MAE), mean squared error (MSE), root mean squared error (RMS) and Pearson's correlation between predicted and ground truth outputs (r). Also reported was the area over the regression error characteristic curve (AOC)~\cite{Bi2003}, which provides an estimate of regression error and is analagous to the area under the receiver-operator characteristic curve for classification. Yoldemir \etal\ didn't perform classification or regression but instead performed unsupervised subnetwork extraction and so, to validate their model, used the Omega index (OmI)~\cite{xie2013overlapping}, which gives a measure of consistency between two sets of graphs~\cite{Yoldemir2015}. Similarly, Yoldemir \etal\ and Arslan \etal\  also measured the consistency of their results using the Dice similarity coefficient (DC), which measures of the degree of overlap between two sets~\cite{Yoldemir2015,arslan2015}.


Finally, in addition to measures which directly report the quality of prediction, Rosa \etal\ reported the sparsity and stability of feature weights learned by the model. These measures are interesting because they give a sense of how consistent the learned model is across training sets and may suggest how well the model will generalize to unseen data.

The papers in Table~\ref{tab:depthReportTable} also used a variety of cross validation schemes to demonstrate the ability of their learning models to generalize to unseen data. Leave-one-out (LOO) cross validation was the most widely used approach, appearing in \nPapersUseLOO\ studies. A variation on this approach, used in particular for cases with imbalanced data sets, is leave-one-subject-per-group-out (LOSGO), which requires at least one instance from each class to be included in every test set, mitigating the bias from class imbalance on the validation measures. Generally, it is possible to perform leave-$p$-out cross validation, in which a random test set of size $p$ is left out from training in every round. This strategy can be useful for reducing the number of times that the a learning model needs to be trained while retaining the same total number of test instances across rounds. Only \nPapersUsedLKO\ machine learning on connectome data studies used leave-$p$-out with $p>1$. However, many studies performed K-fold cross validation in which the entire dataset is split randomly into K equally sized, mutually exclusive subsets (called folds) and the model is trained K times on all but one of the folds and tested on the remaining fold. Often, the random splitting of the data into folds is repeated in multiple rounds and K-fold cross validation is performed once in each round. 

As was mentioned above, different cross validation schemes can yield different results, which is at least in part due to the varying sizes of the training sets used to train the model in each scheme. For instance, Iidaka \etal\ performed 2-fold, 10-fold, 5-fold and leave-one-out cross validation and found wide ranging results (\eg\ from 77\% to 90\% accuracy) across the different schemes~\cite{Iidaka2015}. Beyond yielding varying results, different cross validation schemes will impose different computational loads (\eg\ number of times the learning model must be trained) and yield different statistical powers (\eg\ confidence interval sizes for mean validation measures that are averaged over rounds) based on the number of instances tested and on the number of models trained on unique training subsets. Thus, choice of an appropriate cross-validation scheme depends on the particulars of the data and on the desired subsequent analysis of the results.

\section{Conclusions and Discussion}
\label{sec:conclusions}
Functional and structural connectomes have been proven as informative descriptions of the brain by their use in modelling and classification of a vast array of psychological conditions, neurological conditions, physiological variables and brain states. To do this, researchers have adapted machine learning tools to take advantage of the unique properties of connectome data. Such proposed methods include a variety of local and global connectome features, feature selection and dimensionality reduction techniques that account for the structure and manifold of the connectome data. Additionally, they include graph kernels, topologically aware convolutional layers, objective functions and regularization functions as well as a variety of specialized techniques to extract ROIs and subnetworks from individual connectomes and groups of connectomes. These advances have enabled accurate prediction models as well as techniques for extracting and visualizing salient information from complex brain networks.

Nevertheless, many network specific features and methods for machine learning remain unexplored. For instance, there exist topological measures that, to our knowledge, have never been used in the context of machine learning on connectome data. One such measure is Kirchoff complexity~\cite{tutte2001graph}, which was proposed as a features of structrual connectomes by Li \etal~\cite{Li2013a}. Kirchoff complexity is the count of the number of unique spanning trees in a network and provides a measure of its topology. Note that while Li \etal\ did perform linear regression on these features, they did not split their data into test and training sets and reported results from the context of statistics rather than machine learning (and so this work was not included in Table~\ref{tab:depthReportTable}). Also, while eigenvalues of the Laplacian (\ie\ via spectral decomposition) are fundamental quantities of a network that can be related to almost any other invariant property~\cite{chungspectral}, they remain underutilized, appearing in only one of the papers listed in Table~\ref{tab:depthReportTable}. Such network properties may be important in engineering new connectome features that are sensitive to variations in brain networks indicative of injury and disease.

Alternatively, as the size of connectome datasets increase (see Fig.~\ref{fig:numScansVsYear}), deep learning will become more feasible and will enable models with the flexibility to capture the wide inter-subject variability found in the brain~\cite{Cheng2012}. It seems likely that with more training data, deep models may be able to improve predictive performance and modelling fidelity on connectome data in the same way that it has in other fields (\eg\ natural image classification and natural language processing~\cite{LeCun2015}). One challenge will be to discover how to best leverage the deep learning paradigm for learning on connectome data, given its unique properties. It seems plausible that many of the methods already proposed for learning on connectome data (\eg\ graph kernels and Laplacian based regularization) could be applicable in a deep learning context. Another challenge will be to learn how best to augment datasets. Dataset augmentation is a common practice for deep learning on 2D images, multiplying the size of the training set and allowing known symmetries to be implicitly incorperated into the model~\cite{chatfield2014return}. Standard augmentation operations, including translating and flipping images (encoding invariances to these transforms), are applicable to images of natural scenes but not to connectome data.

Structured prediction is another branch of machine learning that, for the time being, remains under explored on connectome data. Structured prediction models are akin to regressors or classifiers except that, rather than ouputting scalars or vectors of continuous or categorical variables they output structured objects (\eg\ graphs, strings). While none of the papers in Table~\ref{tab:depthReportTable} performed structured prediction, the unique topology of connectome data seems to present interesting possibilties for these models. For instance, indentification of particular subnetworks (\eg\ finding groups of damaged or disease affected connections) or prediction of thoughts or behaviors from functional connectome data might be amenable to a structured prediction formulation.

A variety of papers listed in Table~\ref{tab:depthReportTable} combined structural and functional connectome data and Fig.~\ref{fig:numPapersVsYear} showed that the number of papers using multiple connectome modalities is growing. However, most papers that combined modalities used dMRI only to identify landmarks and studied the fMRI connectivity exclusively. Only a select few studies incorporated both structural and functional connectomes into their learning model but those that did find interesting differences and relationships between the two connectome types that highlight the possible advantages of combining modalities~\cite{Yoldemir2015coupled,Caeyenberghs2013,Wee2012,Chen2015a}. Future work should continue to explore structural and functional connectome data as well as models that combine modalities, especially towards quantifying the strenghts and weaknesses of each modality with respect to modelling different diseases and conditions.

Brain networks that have been studied thus far typically represent connectivity at the largest spatial scale of the brain (\ie\ with each ROI representing millions of neurons). As the spatial, temporal (fMRI) and angular (dMRI) resolutions of scans and scan protocols used to construct connectomes improve, we may choose to construct connectomes that capture these ever finer scales~\cite{sotiropoulos2013advances,uugurbil2013pushing}. Thus, yet another challenge will be to design learning and analysis algorithms that can make use of fine scale information across all or part of these brain networks. 

As the field matures it will be increasingly important establish the robustness of different machine learning methods on larger connectome datasets with large anatomical variation, especially if these methods are to be used in a clinical setting~\cite{walsh2011search}. As mentioned above, dataset sizes are increasing and a handfull of large, open datasets already exist but only for a limited set of neurological conditions. In order to make use of as much data as possible, standards for connectome construction and/or methods which can incorporate connectomes from different construction methods (\eg\ transfer learning~\cite{pan2010survey}), may become important. One step in that direction will be to encourage collaboration and competition in the field, possibly through open challenges with predetermined evaluation approaches and competition leaderboards, which have proved successful in other domains.

We are still, likely, many years away from incorporation of the methods reviewed in this paper as tools into clinical practice. In addition to the lack of large datasets for robust validation and the relatively small number of approaches that have been shown to perform (\ie\ predict) with high accuracies (\eg\ 95-99\%), it is not yet clear exactly how such tools, when they exist, will be used in the clinical workflow. Additionally, machine learning models, which are often `black boxes', face hurdles appealing to regulatory entities (\ie\ FDA) and to clinicians who may wish to understand how a model makes its decisions~\cite{castellanos2013clinical}. Furthermore, not all clinical data is yet structured in a way that makes it ready for computerized analysis (\eg\ qualitiative notes). Nevertheless, as methods continue to improve and data becomes more readily available, machine learning techniques for brain network data remain a promising group of technologies for the future of medicine and neuroscience.

Finally, while we have made every effort to cite all papers related to machine learning on connectome data from MRI, some may have been missed. Furthermore, new papers on this topic will continue to be published as the feild continues to grow. In order to enable our list of studies to become a living document that remains up-to-date, we have created a website at \href{http://connectomelearning.cs.sfu.ca/}{http://connectomelearning.cs.sfu.ca/} with a dynamic version of Table~\ref{tab:depthReportTable} that the community can contribute to. This website will provide researchers with easy access to similar works and better exposure for their new research, perhaps helping to encourage competition and collaboration.

\afterpage{ 
 \newgeometry{left=1cm,right=1cm,top=0.5cm,bottom=0.5cm} 
 \begin{landscape} 
 \centering 
 \scriptsize
\captionof{table}{Comparison between different papers that have applied machine learning to connectome data.} 
\label{tab:depthReportTable}\begin{longtable}{| L{1.7cm} | L{0.9cm} | L{1.6cm} | L{2.8cm} | L{2.6cm} | L{2.7cm} | L{1.8cm} | L{1.9cm} | L{1.5cm} | L{1.3cm} | L{0.6cm} | L{1.3cm} | L{0.9cm} |} 
\hline 
 \textbf{Paper} & \textbf{Image Types} & \textbf{Disease / Groups} & \textbf{Features} & \textbf{Dim. Reduction / Preprocessing} & \textbf{Model / Loss} & \textbf{Regularization} & \textbf{Age Group} & \textbf{\# Scans} & \textbf{ROI Type} & \textbf{\# ROI} & \textbf{Validation} & \textbf{Scheme} \\ 
 \endhead \hline 
\href{http://dx.doi.org/10.1016/j.neuroimage.2015.01.026}{Aln\ae s \etal}, 2015~\cite{Alnæs2015}&fMRI&rest / low / high functional task loads&ROI signal partial + full correlations from GLASSO&-&rLDA&Covariance matrix shrinkage&Adults (29+-10)&101 (37 per class)&Learned (dual-regression)&46&ACC=100 / 70 / 70&LOSGO\\ 
 \hline 
\href{http://dx.doi.org/10.1093/brain/awr263}{Anderson \etal}, 2011~\cite{Anderson2011}&fMRI&Autism&ROI signal correlations&t-test&Age-based per-class linear model fit test&-&Adolecents (8-42)&80 (40+, 40-)&Voxel selection (Min. 5 mm radius)&7266&ACC=79, SN=83, SP=75&LOO\\ 
 \hline 
\href{http://dx.doi.org/10.1016/j.neuroimage.2013.12.015}{Anderson \etal}, 2014~\cite{Anderson2014}&fMRI, Cortical Thickness&ADHD&Local network measures + cortical measurements + metadata&NMF&Decision tree&-&Adolescents ~(8-19)&730 (276+, 472-)&Atlas&90&ACC=67, SN=76, SP=61&LOO\\ 
 \hline 
\href{http://dx.doi.org/10.3389/fnins.2013.00133}{Arbabshirani \etal}, 2013~\cite{Arbabshirani2013}&fMRI&Schizophrenia&ROI signal correlations&Remove features effected by medication, various kernels&Many (KNN, decision trees, RBF neural network, kernel SVM: all tied)&-&Adults (25-50)&56 (28+, 28-)&Learned (ICA)&9&ACC=96, SN=100, SP=92, PPV=92, NPV=100&LOO\\ 
 \hline 
\href{http://dx.doi.org/10.1007/978-3-319-19992-4_7}{Arslan \etal}, 2015~\cite{arslan2015}&fMRI&Healthy Adults&Locally connected cortical ROI signal correlations&-&Multi-graph spectral decomposition&-&Adults (22-35)&40&Learned&50, 100, 150, 200&DC=0.72&leave 20 (50\%) out, 20 rounds\\ 
 \hline 
\href{http://dx.doi.org/10.1109/ICACCI.2015.7275626}{Ashikh \etal}, 2015~\cite{Ashikh2015}&fMRI&NC / eMCI / MCI / AD,&Temporal variance of ROI signal correlations &t-test&OPTICS&-&Seniors (ADNI)&123 (35, 34, 34, 29)&Not specified&200&ACC=100&-\\ 
 \hline 
\href{http://dx.doi.org/10.1016/j.neuroimage.2015.08.055}{Ball \etal}, 2016~\cite{Ball2016}&fMRI&Preterm / term, age&ROI signal covariances&Boruta feature selection&SVM + RBF, Linear regression&-&Preterm (36-49 wks)&131 (105+,26-)&Learned (ICA)&71&BAC=80, AUC=0.92, $r^2$=0.57, MSE=8.9&5 folds\\ 
 \hline 
\href{http://dx.doi.org/10.1016/j.neuroimage.2011.10.002}{Bassett \etal}, 2012~\cite{Bassett2012}&fMRI&Schizophrenia&Largest connected component size curve over thresholds&-&SVM&L2&Adults (30-52)&58 (29+, 29-)&AAL&90&ACC=75, SN=85, SP=64&1000 rounds, leave 28 (14+, 14-) out\\ 
 \hline 
\href{http://dx.doi.org/10.1016/j.neuroimage.2016.09.046}{Brown and Kawahara \etal}, 2016~\cite{kawahara2016}&dMRI&Motor, Cognitive Function, Age&Edge tract counts&-&BrainNetCNN&L2&Preterm&168&Atlas&90&r=0.31, 0.19, 0.86, MAE=10.6, 10.5, 2.3&3 folds\\ 
 \hline 
\href{http://dx.doi.org/10.1007/978-3-319-24553-9_9}{Brown \etal}, 2015~\cite{Brown2015}&dMRI&Low Motor Function&Network Measures&PCA (After net. measures)&SVM&L2&Preterm&168 (22+, 146-)&Atlas&90&ACC=72, SN=77, SP=69&1000 rounds, LOSGO\\ 
 \hline 
\href{http://dx.doi.org/10.1007/978-3-319-46720-7_21}{Brown \etal}, 2016~\cite{Brown2016}&dMRI&Low Motor, Cognitive Function&Edge tract counts&-&Non-negative linear regression&L1, backbone prior, connectivity prior&Preterm&168 (29+, 139-), 168 (13+, 155-) &Atlas&90&ACC=71, 60, AOC=14.3, 17.4, r=0.44, 0.34&1000 rounds, LOSGO\\ 
 \hline 
\href{http://dx.doi.org/10.3389/fnhum.2013.00726}{Caeyenberghs \etal}, 2013~\cite{Caeyenberghs2013}&dMRI + fMRI&TBI&structural and functional nodal degrees&-&Discriminant function analysis&-&Adults (25+/-6)&33 (17+, 16-)&Motor Regions&22&ACC=61, SN=43, SP=77&Not stated\\ 
 \hline 
\href{http://dx.doi.org/10.1148/radiol.10100734}{Chen \etal}, 2011~\cite{chen2011}&fMRI&AD&ROI signal correlations&Wilcoxon rank-sum test&LDA&-&Seniors (69-82)&55 (20NC, 15MCI, 20AD) &AAL&116&AUC=0.87, SN=85, SP=80&LOO\\ 
 \hline 
\href{http://dx.doi.org/10.1007/978-3-319-24571-3_24}{Chen \etal}, 2015~\cite{Chen2015a}&dMRI + fMRI&TBI&Edge FA, ROI signal correlations&-&Multi-view Clustering&-&Not specified&40 (16+, 24-)&DICCCOL&358&-&-\\ 
 \hline 
\href{http://dx.doi.org/10.1109/ISBI.2014.6867977}{Chen \etal}, 2014~\cite{Chen2014a}&dMRI&Healthy Adults&Edge tract counts (binary)&-&Training group-wise consistent connections&-&Not specified&120&DICCCOL, learned&358, 272&-&-\\ 
 \hline 
\href{http://dx.doi.org/10.3389/fnsys.2012.00058}{Cheng \etal}, 2012~\cite{Cheng2012a}&fMRI&ADHD&ROI signal partial + full correlations&t-test&SVM&L2&Adolescents ~(8-19)&730 (276+, 472-)&Atlas&90&ACC=76, SN=63, SP=85&LOO\\ 
 \hline 
\href{http://arxiv.org/abs/1603.06790}{Chung \etal}, 2016~\cite{Chung2016}&dMRI&Low Motor Function&Heat flow features&Flow averaged between 6 partitions&Gaussian naive Bayes&-&Infants (24 months)&290 (55+, 235-)&AAL&116&ACC=82, SN=75, SP=83, fs=79&10 folds, 5 reps\\ 
 \hline 
\href{http://dx.doi.org/10.1002/mrm.22159}{Cradock \etal}, 2009~\cite{Craddock2009}&fMRI&MDD&ROI signal correlations + r-to-z&Reliability filter&SVM&L2&Adults (22-54)&40 (20+, 20-)&AFNI + Relevance to MDD&15&ACC=95, SN=92, SP=96, AUC=0.96&LOO\\ 
 \hline 
\href{http://dx.doi.org/10.1016/j.media.2011.05.007}{Cuingnet \etal}, 2011~\cite{Cuingnet2011}&dMRI&Stroke&Nodal features&-&SVM&Exp. of Laplacian  (over nodes)&Adults (24 - 81)&72&Per-voxel&96 x 64 x 24&ACC=76&Not stated\\ 
 \hline 
\href{http://dx.doi.org/10.1109/ISBI.2015.7163812}{Dodero \etal}, 2015~\cite{Dodero2015}&fMRI, dMRI, fMRI&ASD, ASD, Schizophrenia&Graph Laplacians&-&Kernel SVM&L2&Adolescents, Adults (9-18, 9-18, 33+/-9)&79 (42+, 37-), 94 (51+, 43-), 27 (12+, 15-)&Atlas&264, 264, 74&ACC=61, 74, 68&LOO\\ 
 \hline 
\href{http://dx.doi.org/10.1126/science.1194144}{Dosenbach \etal}, 2011~\cite{Dosenbach2011}&fMRI&Age groups&ROI signal correlations &t-test&SVM&L2&Children / Adults (7-30)&122 (61+, 61-)&Learned&210&ACC=91, SN=90, SP=92&LOO\\ 
 \hline 
\href{http://dx.doi.org/10.1109/PRNI.2014.6858518}{Fei \etal}, 2014~\cite{Fei2014}&fMRI&MCI&gSpan frequent subnetworks from different thresholds&DSM + WL kernel&Graph-kernel-based&-&Seniors (74 +/- 8)&37 (12+, 12-)&AAL&116&ACC=97, AUC=0.96&LOO\\ 
 \hline 
\href{http://dx.doi.og/10.1117/12.2217377}{Galvis \etal}, 2016~\cite{Galvis2016}&dMRI&Parkinson's Disease&ROI signal correlations (no EPI correct, 3DMI, PDEC)&ICC + t-test&SVM&L2&Seniors (51-71)&189 (131+, 58-)&FreeSurfer&68&BAC=60&10 folds, 10 reps\\ 
 \hline 
\href{http://dx.doi.org/10.1007/978-3-319-24571-3_21}{Gao \etal}, 2015~\cite{Gao2015}&dMRI&NC / SMC / MCI / AD&Edge tract counts normalized by ROI volumes&-&Multi-graph normalized-cut&Graph Laplacian (over edges)&Seniors (ADNI)&154 (30 NC, 34 SMC, 62 MCI, 28 AD)&FreeSurfer&129&-&-\\ 
 \hline 
\href{http://hdl.handle.net/10106/25789}{Gellerup \etal}, 2016~\cite{Gellerup2015}&fMRI&Parkinson's Disease&ROI signal correlations + network measures across 5 frequency bands&mRMR&Proximal SVM ensemble&L2&Seniors (60+/-10)&45 (24+, 21-)&Atlas&264&ACC=0.84, SN=0.73, SP=93&5 folds\\ 
 \hline 
\href{http://dx.doi.org/10.1016/j.media.2014.06.006}{Ghanbari \etal}, 2014~\cite{Ghanbari2014}&dMRI&ASD, Age&Edge probability&Built-into NMF&NMF + graph embeddings&-&Adolescents (8-18)&83 (24+, 59-)&Atlas (Desikan)&79&-&-\\ 
 \hline 
\href{http://dx.doi.org/10.1097/WNR.0b013e32835a650c}{Guo \etal}, 2012~\cite{Guo2012}&fMRI&MDD&Nodal degree, efficiency, betweeness centrality from partial correlation at multiple thresholds&-&Many (SVM, RBF-SVM, LDA, random forest, logistic regression)&L2&Adults (17 - 54)&76 (38+,28-)&AAL&90&ACC=79&100 rounds, leave 30\% out\\ 
 \hline 
\href{http://dx.doi.org/10.1016/j.cortex.2014.08.011}{Iidaka}, 2015~\cite{Iidaka2015}&fMRI&ASD&ROI signal correlations &Min. correlation threshold&Probabilistic Neural Networks&-&Children (6-19)&640 (312+, 328-)&AAL&90&ACC=90, SN=92, SP=88, PPV=88, NPV=92, cPPV=8, cNPV=100&2, 10 and 50 folds, Leave one out\\ 
 \hline 
\href{http://dx.doi.org/10.1002/hbm.22353}{Jie \etal}, 2014~\cite{Jie2014}&fMRI&MCI&Local CC from multiple thresholds&t-test, RFE&Multi-kernel SVM&L2&Seniors (74+/-8)&37 (12+, 25-)&AAL&90&ACC=92, BAC=94, SN=100, SP=88, AUC=94&LOO\\ 
 \hline 
\href{http://dx.doi.org.proxy.lib.sfu.ca/10.1109/TBME.2013.2284195}{Jie \etal}, 2013~\cite{Jie2014b}&fMRI&MCI&Local CC from thresholded ROI signal correlations&t-test + RFE + Linear and WL Kernel&Multi-kernel SVM&L2&Seniors (65-83)&37 (12+, 25-)&AAL&116&ACC=92&LOO\\ 
 \hline 
\href{http://dx.doi.org/10.1007/978-3-319-10470-6_90}{Jie \etal}, 2014~\cite{Jie2014a}&fMRI&MCI&CC on Hypergraph&M2TFS + linear kernel&Multi-kernel SVM&L2&Seniors (65-83)&37 (12+, 25-)&AAL&116&ACC=95, SN=92, SP=96, AUC=0.96&LOO\\ 
 \hline 
\href{http://dx.doi.org/10.1007/978-3-319-24888-2_21}{Jin \etal}, 2016~\cite{Jin2015}&dMRI&ASD&Edge FA, MD and TC, Multiple Scales&t-test, LASSO logistic regression&Multi-kernel SVM&L2&Infants&80 (40+, 40-)&Atlas&90, 203, 403&ACC=76, SN=72, SP=79, AUC=0.8&5 folds, 10 reps\\ 
 \hline 
\href{http://doi.org/10.2463/mrms.2015-0027}{Kamiya \etal}, 2016~\cite{Kamiya2015}&dMRI&TLE&Local network measures (4x83=332)&Dantzig selector + RBF Kernel&Kernel SVM&L2&Adults (21-45)&58 (44+, 14-)&FreeSurfer&83&ACC=90, AUC=0.97&LOO\\ 
 \hline 
\href{http://dx.doi.org/10.1016/j.neuroimage.2015.12.028}{Kaufmann \etal}, 2016~\cite{Kaufmann2016}&fMRI&Sleep Deprevation&ROI signal partial correlations from GLASSO&-&rLDA&Covariance matrix shrinkage&Adults (20-24)&60 (41+, 19-)&Learned (MELODIC ICA)&27&ACC=73 / 54 / 85&LOO\\ 
 \hline 
\href{http://dx.doi.org/10.1016/j.clinph.2015.02.060}{Khazaee \etal}, 2015~\cite{Khazaee2015a}&fMRI&AD&Local + global network measures (454)&Fischer score&SVM&L2&Seniors (ADNI)&40 (20+, 20-)&AAL&90&ACC=100&LOO\\ 
 \hline 
\href{http://dx.doi.org/10.1007/s11682-015-9448-7}{Khazaee \etal}, 2015~\cite{Khazaee2015}&fMRI&NC / MCI / AD&Local + global network (2909) on global cost efficiency max. network&Fischer score + FSFS&SVM&L2&Seniors (ADNI ~ 64-83)&168 (41NC, 89MCI, 34AD)&Atlas&264&ACC=88, per-class SN, SP, PPV&Cross-validation with holdout\\ 
 \hline 
\href{http://dx.doi.org/10.1109/ISBI.2012.6235607}{Li \etal}, 2012~\cite{Li2012}&dMRI&ASD&Edge connectivity&RFE&SVM&Graph Laplacian (over edges)&Children (7-14)&20 (10+, 10-)&Atlas&68&ACC=100&LOO\\ 
 \hline 
\href{http://dx.doi.org/10.1002/hbm.22290}{Li \etal}, 2014~\cite{Li2014b}&fMRI + dMRI for landmarks&PTSD&Temporal sliding window nodal degree&Manual selection of quasi-static states&K-means + FDDL, finite state machines&-&Adults&95 (45+, 51-)&DICCCOL&358&Avg. diff. of subnet histograms&10 folds\\ 
 \hline 
\href{http://dx.doi.org/10.1016/j.neuroimage.2016.01.056}{Mitra \etal}, 2016~\cite{Mitra2016}&dMRI&TBI&Edge FA&NBS Edge t-test + PCA&Random Forest&-&Adults&215&AAL&116&ACC=68, SN=80, SP=46, PPV=68, NPV=69&10 folds\\ 
 \hline 
\href{http://dx.doi.org/10.1117/12.911203}{Mokhtari \etal}, 2012~\cite{Mokhtari2012}&fMRI&Rest/Attention&ROI signal correlations&Recursive feature ranking + WL Kernel&Kernel SVM&L2&Adults (20-30)&38 (19+, 19-)&Atlas&24&ACC=100&LOO\\ 
 \hline 
\href{http://dx.doi.org/10.1117/12.2211519}{Moyer \etal}, 2015~\cite{Moyer2015}&dMRI&AD, Bipolar Disorder&Edge tract counts&-&MMSB + (naive Bayes, SVM)&-&ADNI-2, ?&96 (46+, 50-), 92&FreeSurfer&68&ACC=82, 61, REC=80, 53, PRS=82, 69&10 folds\\ 
 \hline 
\href{http://dx.doi.org/10.1016/j.neuroimage.2015.06.008}{Munsell \etal}, 2015~\cite{Munsell2015}&dMRI&TLE, surgical outcome&Edge tract counts&ElasticNet + linear kernel, or SCCA or deep model&SVM&L2&Adults (18 - 70)&118 (70+, 48-)&Atlas (Lusanne)&82&ACC=80, SN=74, SP=84, PPV=90, NPV=70&10 folds\\ 
 \hline 
\href{http://dx.doi.org/10.1007/978-3-319-10470-6_51}{Ng \etal}, 2014~\cite{Ng2016}&fMRI&Before/After Memory Task&Covariance matrices&Matrix whitening transport&SVM&L2&Not specified&102 (51+, 51-)&Atlas&78&ACC=98&10000 rounds, leave 34 out\\ 
 \hline 
\href{http://dx.doi.org/10.1109/PRNI.2012.11}{Ng \etal}, 2012~\cite{Ng2012}&dMRI + fMRI&Viewing a face vs rest&Per - voxel time series.&-&SLR&L1 + Graph Laplacian&Not specified&36 (18+, 18-)&Ward clustering&500&ACC=86&9 folds\\ 
 \hline 
\href{http://dx.doi.org/10.3389/fnhum.2014.00425}{Pariyadath \etal}, 2014~\cite{Pariyadath2014}&fMRI&Smoking Status&Correlations within and between subnetwork regions&RFE&AdaBoost + SVM&L2&Adults (28-50)&42 (21+, 21-)&ICA + Clustering&56&ACC=79, PPV=83&LOO\\ 
 \hline 
\href{http://dx.doi.org/10.1371/journal.pone.0141376}{Park \etal}, 2015~\cite{Park2015}&dMRI, fMRI&BMI&Edge fiber density + fMRI mean nodal degrees&t-test + permutation testing, region prior, functional-structural correlation&PLSR&-&Adults (~29)&120 (60+, 60-)&AAL&116&MAE=15\%, RMS=5.3&LOO\\ 
 \hline 
\href{http://dx.doi.org/10.1109/ISBI.2014.6868000}{Prasad \etal}, 2014~\cite{Prasad2014}&dMRI&AD&Normalized tract counts&Region grouping, PCA&Simulated annealing + SVM&-&Adults (ADNI-2)&87 (37+, 50-)&Freesurfer&68&ACC=85, SN=88, SP=81&LOO\\ 
 \hline 
\href{http://dx.doi.org/10.1016/j.neurobiolaging.2014.04.037}{Prasad \etal}, 2015~\cite{Prasad2015}&dMRI&NC / eMCI / MCI / AD,&Normalized edge TC, max flow between ROIs, network measures&-&SVM&L2&Seniors (64-85)&200 (50, 74, 38, 38)&Freesurfer&68&ACC=78, SN=84, SP=69 &10 folds\\ 
 \hline 
\href{http://dx.doi.org/10.1016/j.dcn.2015.01.003}{Pruett \etal}, 2015~\cite{Pruett2015}&fMRI&Age (6 v.12m) and clincal risk (low, high)&ROI signal correlations&t-test + linear kernel&SVM&L2&Infants (6-12 months)&128 (32 per group)&Atlas&230&ACC=81, 75, SN=78, 81, SP=84, 69&LOO\\ 
 \hline 
\href{http://dx.doi.org/10.1016/j.media.2014.10.006}{Qiu \etal}, 2015~\cite{Qiu2015}&fMRI&Age&ROI signal partial correlations from GLASSO&LLE-SPD&Linear regression&L2&Adults (22-79)&178&Atlas&80&r=0.59, RMS=12.9&LOO\\ 
 \hline 
\href{http://dx.doi.org/10.1016/j.neuroimage.2010.05.081}{Richiardi \etal}, 2011~\cite{Richiardi2011}&fMRI&Resting vs watching movie&Multi-band time series&t-test + FDR&Polythetic trees&Tree pruning + ensembles&Adults (18-36)&15&Altas&90&ACC=97&LOO\\ 
 \hline 
\href{http://dx.doi.org/10.1016/j.neuroimage.2012.05.078}{Richiardi \etal}, 2012~\cite{Richiardi2012}&fMRI&MS&ROI signal correlations&-&Ensemble of functional trees&-&Adults (29-45)&36 (22+, 14-)&AAL&90&ACC=83, SN=82, SP=86&LOO\\ 
 \hline 
\href{http://dx.doi.org/10.1016/j.neuroimage.2010.01.019}{Robinson \etal}, 2010~\cite{Robinson2010}&dMRI&Age (young adults / seniors)&Edge FA&PCA&MLDA&-&Adults (20-30, 59-90)&96&Atlas&83&ACC=87, SN=90, SP=88, Bayes err.=0.87&10 folds\\ 
 \hline 
\href{http://dx.doi.org/10.1016/j.neuroimage.2014.11.021}{Rosa \etal}, 2013~\cite{Rosa2013}&fMRI&MDD&ROI signal sparse inverse covariance&-&SVM&L1&Adults (29-58)&38 (19+, 19-)&Atlas&137&ACC=82, SN=74, SP=89&LOSGO\\ 
 \hline 
\href{http://dx.doi.org/10.1016/j.neuroimage.2014.11.021}{Rosa \etal}, 2015~\cite{Rosa2015}&fMRI&MDD, Depression Spectrum&ROI signal sparse inverse covariance&-&SVM&L1&Adults (29-58), (27-49)&38 (19+, 19-), 60 (30+, 30-)&Atlas&137&ACC=85, SN=83, SP=87, sparsity=0.6\%, stability=57&LOSGO\\ 
 \hline 
\href{http://dx.doi.org/10.3389/fpsyt.2015.00021}{Sacchet \etal}, 2015~\cite{Sacchet2015}&dMRI&MDD&Global Network Measures (9)&-&SVM&L2&Adults (18-55)&32 (14+, 18-)&Atlas&68&ACC=72, SN=71, SP=72&LOO\\ 
 \hline 
\href{http://dx.doi.org/10.1109/ICoBE.2012.6178960}{Shahnazian \etal}, 2012~\cite{Shahnazian2012}&fMRI&Rest / Attention&Bi-variate Granger causality network&Random walk kernel (tried other kernels)&Kernel SVM&L2&Adults (20-30)&38 (19+, 19-)&Atlas&24&ACC=100&LOO\\ 
 \hline 
\href{http://dx.doi.org/10.1016/j.neuroimage.2009.11.011}{Shen \etal}, 2010~\cite{Shen2010}&fMRI&Schizophrenia&ROI signal correlations&KTRCC + LLE &C-Means&-&Adults (19-30)&52 (32+, 20-)&AAL&116&ACC=92, AUC=0.96&LOO\\ 
 \hline 
\href{http://dx.doi.org/10.1016/j.neuroimage.2016.05.029}{Smyser \etal}, 2016~\cite{Smyser2016}&fMRI&Age&ROI signal correlations + r-to-z&t-test + linear kernel&SVM&L2&Preterm Infants (36 - 41 weeks)&100 (50+, 50-)&Spheres in Talairach atlas space&214&ACC=84, SN=90, SP=78&LOO\\ 
 \hline 
\href{http://dx.doi.org/10.1007/978-3-642-40811-3_13}{Sweet \etal}, 2013~\cite{Sweet2013}&fMRI, EEG&Epileptic Regions&ROI signal correlations&-&Baysean model for abnormal regions given edges&-&Not specified&44 (6+, 38-)&Surface subdivision (50-100mm patches)&1153&-&-\\ 
 \hline 
\href{http://dx.doi.org/10.1007/978-3-642-35428-1_23}{Takerkart \etal}, 2012~\cite{Takerkart2012}&fMRI&Differing Auditory Stimuli&ROI time series, ROI barycenters&Functional, geometrical, structural kernels&Kernel SVM&L2&Not specified&45 (9 per class)&Clustering (Ward's)&5-30&ACC=45&LOO\\ 
 \hline 
\href{http://dx.doi.org/10.1098/rstb.2015.0111}{Tunc \etal}, 2016~\cite{Solmaz2016}&dMRI&Sex&Mean anatomical subnetwork inter-connectivity, cognitive + motor scores&-&SVM&L2&Adolescents (15+/-3.5)&900&Atlas (Desikan)&95&ACC=79, 64&10 folds\\ 
 \hline 
\href{http://dx.doi.org/10.1007/978-3-319-24888-2_28}{Vanderweyen \etal}, 2015~\cite{vanderweyen2013}&fMRI&TBI and AD&ROI signal partial correlations&LASSO&SVM&L2&Seniors (~70)&69 (40NC, 15AD, 14TBI)&Atlas&264&ACC=82, SN=40, SP=98, PPV=86, NPV=81&LOO\\ 
 \hline 
\href{http://dx.doi.org/10.1007/978-3-642-15705-9_25}{Varoquax \etal}, 2010~\cite{Varoquaux2010}&fMRI&Stroke&ROI signal correlations&-&Gaussian model over manifold of (control) covariance matrices&-&Not specified&30 (10+, 20-)&Seeded Regions&33&Class log likelihood&LOO\\ 
 \hline 
\href{http://dx.doi.org/10.1109/PRNI.2014.6858549}{Vegas-Pons \etal}, 2014~\cite{Vega-Pons2014}&fMRI&Differing Auditory Stimuli&Thresholded ROI signal correlations&WL kernel&Kernel SVM&-&Not specified&38&Clustering&?&ACC=74&LOO\\ 
 \hline 
\href{http://dx.doi.org/10.1109/ICDMW.2014.98}{Wang \etal}, 2014~\cite{Wang2014}&fMRI&MCI&Local CC from thresholded ROI signal correlations&t-test + RFE + gSpan + Linear, WL Kernels&Multi-kernel SVM&L2&Seniors (65-83)&37 (12+, 25-)&AAL&116&ACC=97, AUC=0.92&LOO\\ 
 \hline 
\href{http://dx.doi.org/10.1111/cns.12499}{Wee \etal}, 2016~\cite{Wee2016}&fMRI&ASD&ROI signal correlations from temporal clusters&LASSO + various kernels&Kernel SVM&L2&Children, Young Adults (4 - 22)&92 (45+, 47-)&AAL&116&ACC=71, SN=80, SP=61, PPV=79, NPV=65&10 folds\\ 
 \hline 
\href{http://dx.doi.org/10.1016/j.neuroimage.2010.10.026}{Wee \etal}, 2010~\cite{Wee2011}&dMRI&MCI&Local CC (for 6 connectome types)&Ranked correlation + RFE&SVM&L2&Seniors (74+/-8.6)&27 (10+,17-)&AAL&90&ACC=89, AUC=0.93&LOO\\ 
 \hline 
\href{http://dx.doi.org/10.1007/978-3-319-02267-3_18}{Wee \etal}, 2013~\cite{Wee2013}&fMRI&MCI&Temporal sliding window region activations&Fused multiple graphical LASSO + t-test&SVM&L2&ADNI-2 (68-80)&59 (29+, 30-)&AAL&116&ACC=90, BAC=79, SN=76, SP=83, AUC=0.79&LOO\\ 
 \hline 
\href{http://dx.doi.org/10.1007/978-3-642-40763-5_40}{Wee \etal}, 2013~\cite{Wee2013a}&fMRI&AD, MCI&Granger causality networks&t-test on MAR model order distribution + RFE&SVM&L2&Seniors (65-83)&37 (12+, 25-)&AAL&116&ACC=92, AUC=90&LOO\\ 
 \hline 
\href{http://dx.doi.org/10.1016/j.neuroimage.2011.10.015}{Wee \etal}, 2012~\cite{Wee2012}&dMRI + fMRI&MCI&Structural and functional nodal CC&t-test, Linear + Polynomial + RBF kernels&Multi-kernel SVM&L2&Seniors (64-83)&27 (10+, 17-)&AAL&90&ACC=96, SN=100, SP=94, AUC=95, YDI=94, BAC=95, FS=97&LOO\\ 
 \hline 
\href{http://dx.doi.org/10.1007/978-3-319-19992-4_61}{Yoldemir \etal}, 2015~\cite{Yoldemir2015coupled}&dMRI + fMRI&7 Functional Tasks&fMRI time series, weighted nodal degree&-&CSORD + SVM&Non-negative weights, fixed weight magnitude&Adults&40&Clustering&200&ACC=79, ICC=0.43, DC=0.63&100 rounds, leave 5 (12.5\%) out\\ 
 \hline 
\href{http://dx.doi.org/10.1109/TMI.2015.2480864}{Yoldemir \etal}, 2013~\cite{Yoldemir2015}&fMRI&Healthy Adults&fMRI time series&-&SORD&Non-negative weights, fixed weight magnitude&Adults&40&Clustering&200&ICC of net. measures, OmI=0.72 &Test-retest\\ 
 \hline 
\href{http://dx.doi.org/10.3389/fnins.2015.00257}{Zhan \etal}, 2015~\cite{Zhan2015}&dMRI + MRI&NC / MCI / AD&Edge tract counts&High order SVD&SLR&L1&ADNI-2&202&Atlas&113&ACC=71, SN=68, SP=72, AUC=0.76&20 rounds, leave 15\% out\\ 
 \hline 
\href{http://dx.doi.org/10.1109/ISBI.2014.6867874}{Zhu \etal}, 2014~\cite{Zhu2014}&fMRI, dMRI for landmarks&MCI, Schizophrenia&ROI signal correlations&Edge t-test + CFS&SVM&L2&Adults&28 (10+,18-), 20 (10+,10-)&DICCCOL&358&ACC=96, 100&LOO\\ 
 \hline 
\href{http://dx.doi.org/10.1002/hbm.22373}{Zhu \etal}, 2014~\cite{Zhu2014a}&fMRI + dMRI for landmarks&MCI&ROI signal correlations&CFS&SVM&L2&Seniors (55-84), (66-84)&28 (10+, 18-), 24 (10+, 10-)&DICCCOL&358&ACC=100, 96&LOO\\ 
 \hline 
\href{http://dx.doi.org/10.1371/journal.pone.0078824}{Ziv \etal}, 2013~\cite{Ziv2013}&dMRI&Neonatal Encephalopathy&Counts of binary subgraphs&RFE, PCA&SVM&L2&Neonates&24&Clustering (Recursive partitioning)&100&ACC=79&Cross-validation\\ 
 \hline 
\end{longtable} 
 \end{landscape} 
 \restoregeometry 
 } 
 
\restoregeometry 

\bibliographystyle{abbrv}
\bibliography{DepthReport}{}


\end{document}